\documentclass{article}

\usepackage{microtype}
\usepackage{graphicx}
\usepackage{subcaption}
\usepackage{booktabs}

\usepackage{hyperref}
\usepackage{enumitem}
\usepackage[table]{xcolor}
\newcommand{\gc}{\cellcolor{yellow!15}}
\usepackage{CJKutf8}

\usepackage[preprint]{icml2026}

\usepackage{amsmath}
\usepackage{amssymb}
\usepackage{mathtools}
\usepackage{amsthm}

\usepackage{multirow}
\usepackage{makecell}
\usepackage{pifont}

\newcommand{\xmark}{\ding{55}}

\usepackage[capitalize,noabbrev]{cleveref}

\theoremstyle{plain}

\theoremstyle{definition}

\theoremstyle{remark}

\usepackage[textsize=small]{todonotes}
\begin{document}

\twocolumn[
  \icmltitle{Tail-Aware Post-Training Quantization for 3D Geometry Models}

  \icmlsetsymbol{equal}{*}

  \begin{icmlauthorlist}

    \icmlauthor{Sicheng Pan}{equal,add0}
    \icmlauthor{Chen Tang}{equal,add1}
    \icmlauthor{Shuzhao Xie}{equal,add0}
    \icmlauthor{Ke Yang}{add2}
    \icmlauthor{Weixiang Zhang}{add0} 
    \icmlauthor{Jiawei Li}{add3} \\
    \icmlauthor{Bin Chen}{add2}
    \icmlauthor{Shu-Tao Xia}{add0}
    \icmlauthor{Zhi Wang}{add0}

  \end{icmlauthorlist}

  \icmlaffiliation{add0}{Tsinghua Shenzhen International Graduate School, Tsinghua University, Shenzhen, China}
  \icmlaffiliation{add1}{MMLab, The Chinese University of Hong Kong, Hong Kong, China}
  \icmlaffiliation{add2}{Harbin Institute of Technology Shenzhen, China}
  \icmlaffiliation{add3}{Huawei Technology, Shenzhen, China}

  \icmlcorrespondingauthor{Bin Chen}{chenbin2021@hit.edu.cn}
  \icmlcorrespondingauthor{Zhi Wang}{wangzhi@sz.tsinghua.edu.cn}

  \icmlkeywords{Machine Learning, ICML}

  \vskip 0.3in
]

\printAffiliationsAndNotice{\icmlEqualContribution}

\begin{abstract}

The burgeoning complexity and scale of 3D geometry models pose significant challenges for deployment on resource-constrained platforms.
While Post-Training Quantization (PTQ) enables efficient inference without retraining, conventional methods, primarily optimized for 2D Vision Transformers, fail to transfer effectively to 3D models due to intricate feature distributions and prohibitive calibration overhead.
To address these challenges, we propose \textbf{TAPTQ}, a Tail-Aware Post-Training Quantization pipeline specifically engineered for 3D geometric learning. Our contribution is threefold: 
(1) To overcome the data-scale bottleneck in 3D datasets, we develop a progressive coarse-to-fine calibration construction strategy that constructs a highly compact subset to achieve both statistical purity and geometric representativeness. 
(2) We reformulate the quantization interval search as an optimization problem and introduce a ternary-search-based solver, reducing the computational complexity from $\mathcal{O}(N)$ to $\mathcal{O}(\log N)$ for accelerated deployment. 
(3) To mitigate quantization error accumulation, we propose TRE-Guided Module-wise Compensation, which utilizes a Tail Relative Error (TRE) metric to adaptively identify and rectify distortions in modules sensitive to long-tailed activation outliers. 
Extensive experiments on the VGGT and Pi3 benchmarks demonstrate that TAPTQ consistently outperforms state-of-the-art PTQ methods in accuracy while significantly reducing calibration time. The code will be released soon.
\end{abstract}

\enlargethispage{3\baselineskip}

\section{Introduction}
\begin{figure*}[t]
    \centering
    \includegraphics[width=\textwidth]{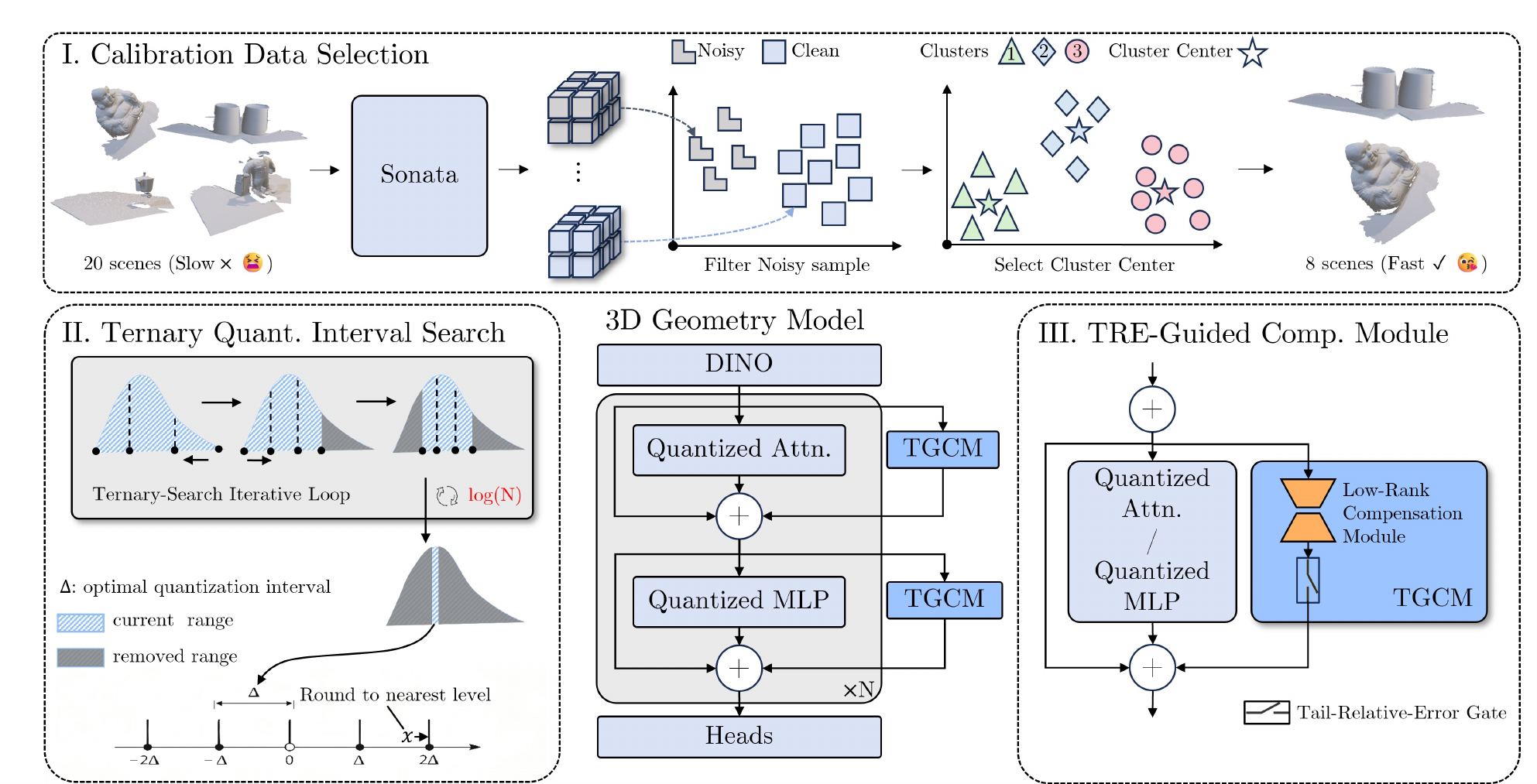}
    \caption{
    \textbf{Overview of TAPTQ.}
    \textbf{(I)} Our proposed Representative Calibration Data via Two-Stage Cluster.
    \textbf{(II)} Our proposed Quantization Interval Optimization via Ternary
Search.
    \textbf{(III)} Our proposed Tail-Relative-Error (TRE)-Guided Compensation Module.
    }
    \label{fig:pipeline}
\end{figure*}
The emergence of large-scale 3D foundation models~\cite{wang2024dust3r,leroy2024grounding,wang2025vggt,lin2025depth} has revolutionized the field of computer vision by enabling end-to-end geometric learning directly from massive datasets. 
This paradigm shift addresses the fundamental challenge of inferring 3D structures from 2D images, which is a cornerstone task that has traditionally relied on Structure-from-Motion (SfM) pipelines~\cite{cui2017hsfm,schonberger2016structure}. 
By employing unified, feed-forward architectures, these modern models jointly solve diverse tasks like multi-view reconstruction and camera estimation with unprecedented accuracy. 
These advances hold immense potential for real-world applications in augmented reality (AR), autonomous driving, and robotics. 
However, the superior performance of these models is inextricably tied to their burgeoning scale, which results in massive computational and memory footprints. 
This presents significant hurdles for practical deployment on resource-constrained or latency-sensitive edge platforms. 

Post-training Quantization (PTQ)~\cite{wei2022qdrop, frantar2022gptq, yuan2022ptq4vit} has emerged as a pivotal technique for model compression, facilitating the conversion of high-precision weights and activations into low-bit integer arithmetic without exhaustive retraining. While PTQ has achieved remarkable success in Large Language Models (LLMs) and 2D Vision Transformers (ViTs), a systematic investigation into its efficacy for 3D geometry foundation models remains largely underexplored. 
Our experimental results show that directly transplanting established PTQ recipes from the 2D vision to 3D geometry transformers leads to severe performance degradation and prohibitive calibration overhead. 
Such failures suggest that the unique properties and multi-view dependencies of 3D geometric data require a specialized quantization paradigm that goes beyond existing general-purpose frameworks.

We attribute these observed failures to the inherent differences of both 3D geometric data and model, which manifest as three primary obstacles: 
\textbf{First, statistical heterogeneity in multi-view calibration.} 
Unlike 2D image tasks, a single calibration instance in 3D geometry models typically comprises multiple correlated yet non-identical views. The resulting activation distributions are highly sensitive to viewpoint configurations and noisy geometric structures, leading to substantial inter-sample statistical variance. Consequently, a sparse calibration set may fail to encapsulate the underlying data manifold, causing range estimation to be biased by unstable outliers. While recent work~\cite{feng2025quantized} attempts to mitigate this through model-dependent filtering, such approaches often necessitate auxiliary forward passes and lack cross-architecture portability. 
\textbf{Second, the computational bottleneck of range optimization.} High-fidelity PTQ relies on optimizing clipping ranges or scaling factors via iterative evaluations. In 3D geometry settings, this cost is severely exacerbated by the multi-view inference paradigm and massive token sequences (e.g., dense point maps), rendering standard $\mathcal{O}$(N) grid searches or reconstruction-based optimizations (e.g., BRECQ~\cite{li2021brecq}) computationally prohibitive. This necessitates a more efficient, ideally logarithmic-complexity search paradigm to accelerate calibration without sacrificing quantization fidelity.
\textbf{Third, intensified error sensitivity in multi-task geometric inference.} 
Unlike previous architecture (e.g., 2D classification via ViTs), which exhibits higher tolerance to quantization noise, 3D geometry foundation naturally models demand significantly more stringent numerical precision . 
In complex tasks such as camera pose estimation and dense reconstruction, even minor quantization errors can amplify across transformer layers, leading to significant geometric drift and misalignment. 
This necessitates a principled and lightweight metric to selectively identify and rectify ``bottleneck'' modules to ensure geometric fidelity with minimal compensation overhead. 

To address these challenges, we propose \textbf{TAPTQ} (\textbf{T}ail-\textbf{A}ware \textbf{P}ost-\textbf{T}raining \textbf{Q}uantization), a high-efficiency pipeline specifically tailored for 3D geometry models. TAPTQ is built upon three core innovations:
\textbf{(i) Progressive calibration set construction:} To ensure robustness under constrained data budgets, we develop a progressive coarse-to-fine calibration construction strategy to construct a compact yet statistically representative calibration set. This strategy suppresses the influence of unstable samples and noisy geometric structures, providing a solid foundation for consistent range estimation.
\textbf{(ii) Ternary-search-based interval optimization:} We reformulate quantization interval selection as an optimization problem. By exploiting the near-unimodal property of the calibration objective, we introduce a ternary-search-based solver that reduces the search complexity from $\mathcal{O}(N)$ to $\mathcal{O}(\log N)$, significantly accelerating the deployment process without compromising quantization fidelity.
\textbf{(iii) TRE-Guided Module-wise Compensation:} To mitigate error accumulation across transformer layers, we propose a selective compensation mechanism. We introduce the Tail Relative Error (TRE) metric to quantify distortions in high-magnitude activation regions. TRE acts as a gating criterion to adaptively identify and rectify only the most affected ``bottleneck'' modules, preserving geometric consistency with minimal computational overhead.

Our contributions are summarized as follows:
\begin{itemize}
    \item We analyze why existing PTQ pipelines for LLMs and 2D vision transformers do not directly translate to 3D geometry transformers, emphasizing calibration fragility, interval-search cost, and the need for selective error correction.
    \item We adopt a progressive coarse-to-fine calibration construction strategy that constructs a compact yet robust calibration set under limited calibration budgets.
    \item We introduce a ternary-search-based interval calibration method that reduces the interval-search complexity from $O(N)$ to $O(\log N)$.
    \item We propose TRE-Guided Module-wise Compensation, using Tail Relative Error to selectively activate compensation and mitigate cross-layer error accumulation with minimal overhead.
\end{itemize}

\section{Related Work}
\subsection{Feed-forward 3D Geometry Models}
Traditional 3D reconstruction typically follows a two-stage paradigm: Structure-from-Motion (SfM) for pose estimation followed by Multi-View Stereo (MVS) for dense geometry \citep{schonberger2016structure, yao2018mvsnet}. While neural radiance fields (NeRF) \citep{mildenhall2021nerf, barron2021mip} and 3D Gaussian Splatting \citep{kerbl20233d,xie2024mesongs,xie2024sizegs} have revolutionized high-fidelity rendering, they fundamentally rely on \emph{per-scene optimization}, which is computationally prohibitive for large-scale, real-time applications.
To circumvent this, recent research has shifted toward \emph{feed-forward geometry foundation models}. DUSt3R \citep{wang2024dust3r} and MASt3R \citep{leroy2024grounding} pioneered direct geometry inference from unconstrained image collections by predicting dense pointmaps. VGGT \citep{wang2025vggt} further unified multiple geometric tasks within a monolithic transformer architecture. Subsequent works have explored architectural efficiency through permutation-equivariant designs \citep{wang2025pi}, unified depth-ray prediction\citep{lin2025depth}, and attention sparsification \citep{wang2025faster, shen2025fastvggt}. Despite these algorithmic advances, the massive parameter count and memory footprint of these transformers remain a bottleneck for edge deployment, necessitating effective model compression.
\subsection{Post-Training Quantization (PTQ)}
PTQ enables model compression using limited calibration data without the need for high-cost quantization-aware training (QAT) \citep{gholami2022survey}. Early methods focused on MSE-driven range clipping \citep{banner2019post} or weight equalization \citep{nagel2019data}. To achieve lower bit-widths, reconstruction-based approaches like AdaRound \citep{nagel2020up}, BRECQ \citep{li2021brecq}, and QDrop \citep{wei2022qdrop} optimize rounding or local block-wise behavior to minimize output distortion.
The shift toward Transformer architectures introduced new challenges, specifically regarding non-linearities (Softmax/GELU) and heavy-tailed activation distributions. In the vision domain, methods such as PTQ4ViT \citep{yuan2022ptq4vit} and RepQ-ViT \citep{li2023repq} address these through specialized scaling and re-parameterization. Similarly, for Large Language Models (LLMs), frameworks like GPTQ \citep{frantar2022gptq}, SmoothQuant \citep{xiao2023smoothquant}, and AWQ \citep{lin2024awq} leverage one-shot weight updates or activation migration to preserve model capacity. However, recent analysis \citep{he2025preserving} suggests that the efficacy of these methods is highly sensitive to the choice of calibration data.
Quantizing large-scale 3D geometry transformers is an emergent and under-explored area. While QuantVGGT \citep{feng2025quantized} provided a preliminary study on the sensitivity of 3D activations, it does not fully address the calibration inefficiency and the complex error accumulation inherent in multi-view geometry tasks. Unlike general-purpose PTQ for 2D ViTs or LLMs, our work explicitly targets the unique failure modes of 3D models—specifically the tail-dominated activation ranges and view-dependent calibration variability. By introducing a ternary-search-based optimization and a tail-relative error metric, we bridge the gap between high-performance geometry foundation models and efficient hardware deployment.

\section{Method}

\subsection{Preliminary}
\label{sec:preliminary}
Post-training quantization (PTQ) compresses a pretrained model into low-bit arithmetic without retraining, 
by quantizing weights and activations using a small unlabeled calibration set. 
In this work, we adopt twin uniform quantization~\cite{yuan2022ptq4vit}, which maps a real-valued tensor $x$ to its quantized version $x_q$ via
\begin{equation}
x_q=
\begin{cases}
\mathrm{clip}\!\left(\left\lfloor \dfrac{x}{\Delta_{R_1}} \right\rceil, q_{\min}, q_{\max}\right)\cdot \Delta_{R_1},
& x \in \mathcal{R}_1, \\[6pt]
\mathrm{clip}\!\left(\left\lfloor \dfrac{x}{\Delta_{R_2}} \right\rceil, q_{\min}, q_{\max}\right)\cdot \Delta_{R_2},
& x \in \mathcal{R}_2 .
\end{cases}
\end{equation}
where $\Delta>0$ denotes the quantization interval (i.e., scaling factor) and
$[q_{\min}, q_{\max}]$ are quantization ranges that are determined by the given bit-width setting.

The central goal of PTQ is to determine $\Delta$, because a small interval leads to saturation while a large interval increases rounding error~\cite{tang2022mixed}. 
PTQ therefore determines $\Delta$ by comparing the outputs of the quantized model with those of the full-precision model 
on calibration data. 
Given the calibration inputs $\mathbf{x}$ and the model $\mathcal{F}(\cdot)$, let $\mathbf{y}_\text{FP}=\mathcal{F}(\mathbf{x})$ be the full-precision output and
$\mathbf{y}_{\Delta}=\mathcal{F}(\mathbf{x};\Delta)$ the corresponding quantized output using the quantization intervals $\Delta$. 
Quantization interval optimization can be formulated as
\begin{equation}
\alpha^{*} = \arg\max_{\alpha\in\mathcal{A}} \ \mathrm{Sim}\!\left(\mathbf{y}_\text{FP}, \mathbf{y}_{\alpha}\right),
\end{equation}
where $\mathcal{A}=\{\alpha_i\}_{i=1}^{N}$ is a discrete set of candidate intervals and
$\mathrm{Sim}(\cdot)$ denotes a similarity metric, defined as
\begin{equation}
\mathrm{Sim}(\mathbf{y}_\text{FP}, \mathbf{y}_{\alpha})
= - \left\lVert \mathbf{g} \odot \left(\mathbf{y}_\text{FP} - \mathbf{y}_{\alpha}\right) \right\rVert_2^2 ,
\end{equation}
where $\mathbf{g}$ is the gradient associated with the corresponding tensor, weighting quantization errors by local sensitivity.

Previous works~\cite{yuan2022ptq4vit,li2023q} evaluate all candidates in $\mathcal{A}$ and thus incur $\mathcal{O}(N)$ forward passes, 
which becomes a major calibration bottleneck for large 3D geometry models. 
This motivates efficient interval search strategies for PTQ, which we address in the following section. 

\subsection{Progressive Calibration Set Construction}

\begin{figure}[t]
  \centering
  \includegraphics[width=0.9\linewidth]{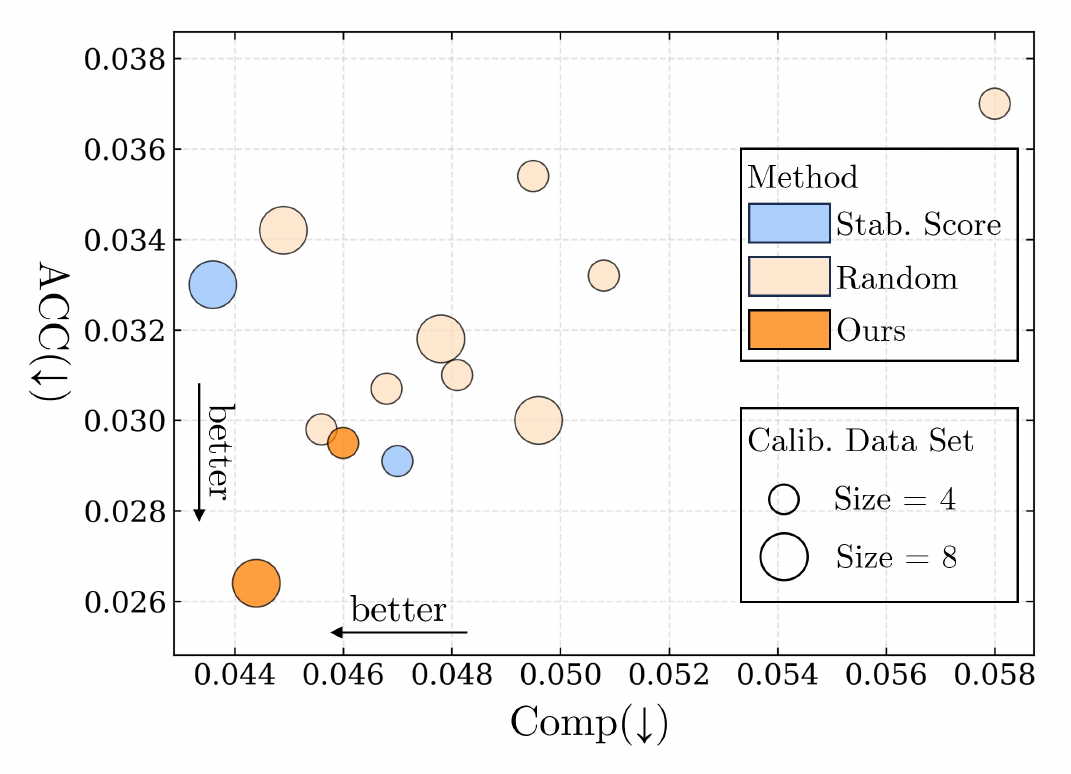}
    \caption{
        \textbf{Acc. and Comp. performance under different calibration data set.}
        Each bubble corresponds to one method configuration on 7Scenes (W4A8),
        where bubble size indicates the number of calibration samples (4 / 8 / 20 scans).
        \emph{Stab. score} denotes a stability-based calibration sample selection strategy (see Appendix~\ref{appendix:stability}).
    }
  \label{fig:calibration_variance}
\end{figure}

Post-training quantization relies heavily on the quality of the calibration set to estimate accurate quantization ranges. Conventional PTQ frameworks typically resort to random sampling~\cite{wei2022qdrop, frantar2022gptq}; however, we observe that for 3D geometry foundation models, such a stochastic approach leads to significant performance fluctuations. As demonstrated in Fig.~\ref{fig:calibration_variance}, repeated random samplings result in a high variance of quantization accuracy, indicating that \textit{unstable or noisy 3D samples can disproportionately distort the estimated quantization ranges}. To address this, we propose a progressive coarse-to-fine calibration construction strategy to achieve both \textit{statistical purity} and \textit{geometric representativeness}. 

\textbf{Stage 1: Coarse-grained Outlier Suppression.} 
The initial stage aims to purify the calibration pool by filtering out noisy instances. Given an unlabeled dataset $\mathcal{D} = \{x_i\}_{i=1}^M$, we first map each 3D sample $x_i$ to a latent feature space $\mathcal{Z}$ using a self-supervised point cloud encoder $f_{\theta}(\cdot)$, such that $z_i = f_{\theta}(x_i) \in \mathbb{R}^d$. We perform $K$-means clustering on $\mathcal{Z}$ with $K=2$ to partition the dataset into two clusters, $C_1$ and $C_2$. 

Guided by the Pareto Principle in large-scale datasets, we hypothesize that high-quality samples constitute the statistical majority, whereas noisy or degenerate samples reside in a sparse minority cluster. We select the purified candidate set $\mathcal{D}_{pure}$ based on a majority-voting mechanism:
\begin{equation}
    \mathcal{D}_{pure} = \text{arg max}_{C \in \{C_1, C_2\}} |C|,
\end{equation}
where $|C|$ denotes the cardinality of the cluster. This procedure effectively suppresses the influence of unstable outliers, providing a robust distribution for subsequent selection.

\textbf{Stage 2: Fine-grained Geometric Representation.} 
After denoising, we seek to construct a final calibration set $\mathcal{C}$ ($|\mathcal{C}|=N$) that captures both diverse geometric configurations and local statistical variations. We perform fine-grained clustering on the features of $\mathcal{D}_{pure}$ with $K = N/2$ clusters, denoted as $\{S_k\}_{k=1}^{N/2}$. For each cluster $S_k$ with centroid $\mu_k$, we select two representative samples closest to the cluster centroid, defined as: 
\begin{equation}
    z_{k,1} = \text{arg min}_{z \in S_k} \| z - \mu_k \|_2.
\end{equation}
\begin{equation}
    z_{k,2} = \text{arg min}_{z \in S_k \setminus \{z_{k,1}\}} \| z - \mu_k \|_2.
\end{equation}

The final calibration set is defined as $\mathcal{C} = \bigcup_{k=1}^{N/2} \{z_{k,1}, z_{k,2}\}$. This strategy ensures that $\mathcal{C}$ not only spans the broad geometric manifold but also encapsulates local variations, thereby enhancing the stability of the estimated quantization parameters.

\begin{figure}[t]
    \centering
    \includegraphics[width=1\linewidth]{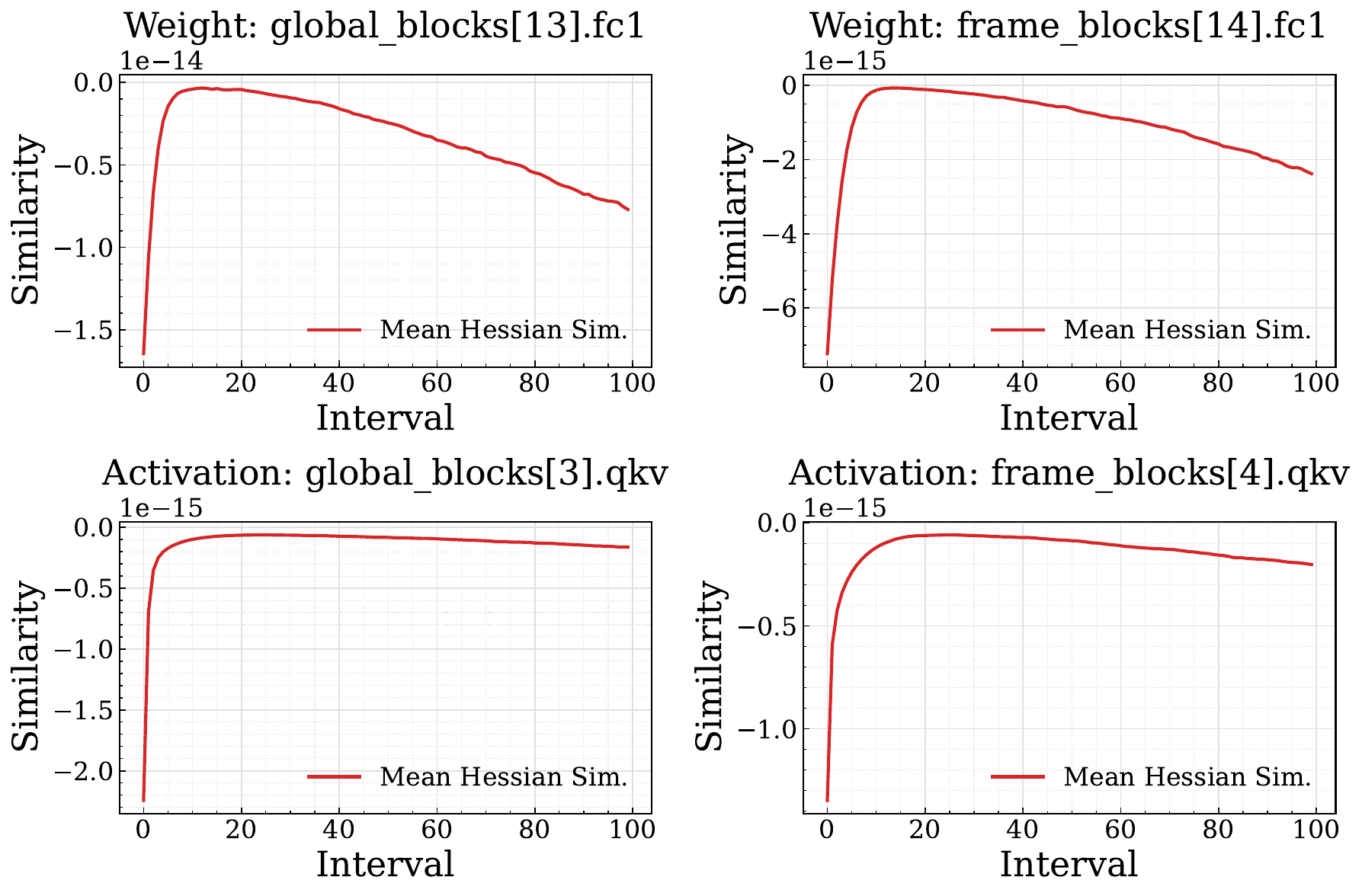}

  \caption{\textbf{Unimodal behavior of similarity with respect to quantization interval.}
  The similarity between quantized and full-precision outputs is evaluated over a discrete
  set of interval candidates for both activations and weights.
  In both cases, the similarity exhibits an approximately unimodal trend, enabling efficient
  interval search via ternary search.}
  \label{fig:interval_unimodal}
\end{figure}

\subsection{Quantization Interval Optimization via Ternary Search}

In conventional PTQ frameworks, interval selection is typically treated as an exhaustive grid search over a discrete candidate set $\mathcal{A}=\{\alpha_i\}_{i=1}^{N}$. 
This incurs prohibitive $\mathcal{O}(N)$ computational overhead, as each candidate requires a complete quantized forward pass.
For large-scale 3D geometry models, this linear complexity becomes a critical calibration bottleneck, with execution times often exceeding 522.6 GPU-minutes per model. 
To alleviate this, we propose an accelerated interval optimization strategy based on the \textit{ unimodality} of the quantization objective.

\textbf{Unimodality of the Similarity Landscape.} 
We investigate the behavior of the similarity function $\mathrm{Sim}(\mathbf{y}_\text{FP}, \mathbf{y}_{\alpha})$ across the candidate space $\mathcal{A}$. As illustrated in Fig.~\ref{fig:interval_unimodal}, the similarity score exhibits a consistent \textit{near-unimodal} trend for both weights and activations. 
This phenomenon is rooted in the fundamental trade-off of quantization: a narrow interval leads to excessive \textit{clipping distortion}, while an overly wide interval increases \textit{rounding noise} by coarsening the quantization step size. 
The competition between these two error sources naturally induces a unimodal landscape, where a unique global optimum exists.

\textbf{Ternary Search Solver.} 
Leveraging the unimodal property, we reformulate interval selection as a constrained optimization problem. Instead of a linear scan, we employ a ternary search algorithm to progressively narrow the search boundaries $[L, R]$ within the discrete index range $[1, N]$. At each iteration, we evaluate two interior points $m_1$ and $m_2$:
\begin{equation}
    m_1 = L + \frac{R-L}{3}, \quad m_2 = R - \frac{R-L}{3}.
\end{equation}
By comparing the similarity scores $\mathrm{Sim}_{m_1}$ and $\mathrm{Sim}_{m_2}$ obtained from the quantized forward passes, the search space is updated as follows:
\begin{equation}
    [L, R] \leftarrow 
    \begin{cases} 
    [m_1, R] & \text{if } \mathrm{Sim}_{m_1} < \mathrm{Sim}_{m_2}, \\
    [L, m_2] & \text{otherwise.}
    \end{cases}
\end{equation}
The search terminates when $R - L \leq \epsilon$, where $\epsilon$ is a small convergence threshold (e.g., $\epsilon=0.0001$). This strategy is applicable to both weight and activation calibration, ensuring broad portability across different layers.

\textbf{Complexity Analysis.} 
While conventional linear search requires $\mathcal{O}(N)$ evaluations, our ternary search solver reduces the complexity to $\mathcal{O}(\log N)$. In practice, this logarithmic reduction transforms the calibration process from a time-intensive bottleneck into a near-instantaneous operation. Furthermore, our method avoids the sub-optimal local minima often inherent in coarse grid searches, maintaining high quantization fidelity while significantly accelerating the deployment pipeline.

\begin{figure}[t]
  \centering
  \includegraphics[width=\columnwidth]{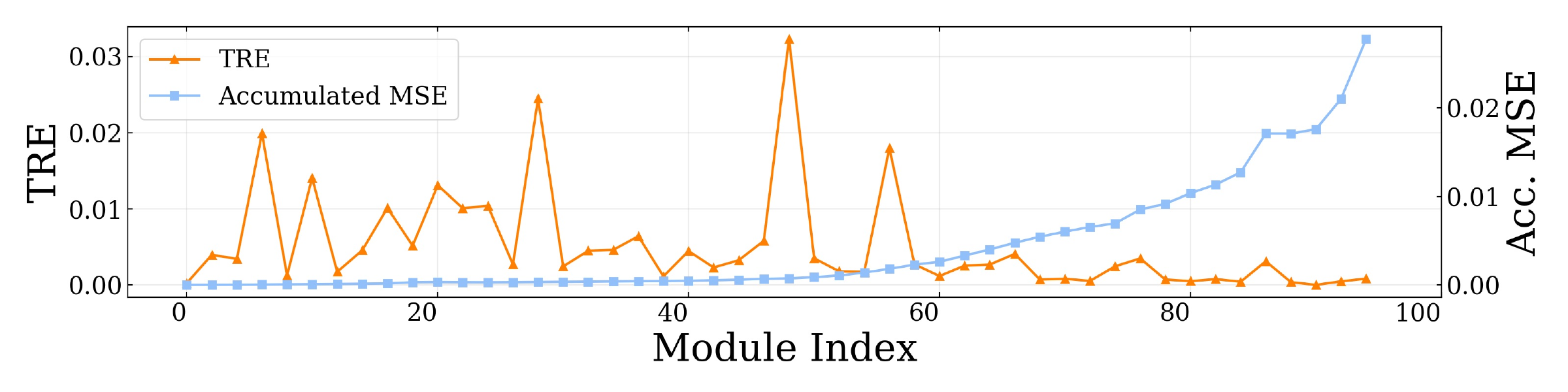}
    \caption{\textbf{Accumulated MSE and Tail Relative Error across W4A8 quantized Transformer modules.}
    The accumulated MSE is measured at the output of each quantized module along the network depth.
    Results are reported for a quantized VGGT model with layer-wise calibration applied.}
  \label{fig:state_error}
\end{figure}

\begin{figure}[t]
  \centering
  \includegraphics[width=\columnwidth]{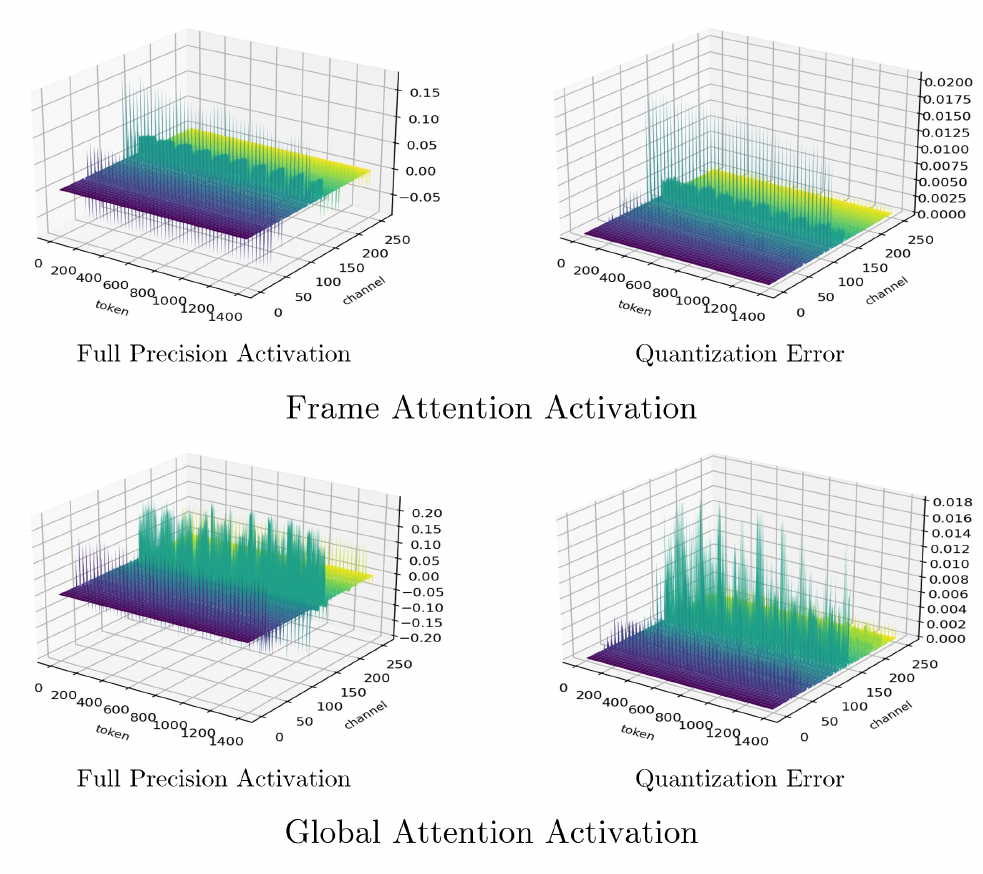}
  \caption{\textbf{Visualization of activation values and corresponding quantization errors.}
    Each horizontal pair shows the full-precision activations (left) and the corresponding quantization errors (right) for representative attention modules. Both distributions exhibit heavy-tailed characteristics, with large-magnitude activations contributing disproportionately to the observed quantization error.}
  \label{fig:activation_error_tail}
\end{figure}

\subsection{TRE-Guided Module-wise Compensation}

Despite optimal interval selection, PTQ for deep 3D geometry transformers still suffers from \textit{cascading quantization noise}. 
Since PTQ follows a sequential calibration paradigm without end-to-end optimization, quantization errors introduced in preceding layers inevitably accumulate and amplify as they propagate through the network hierarchy. 
As illustrated in Fig.~\ref{fig:state_error}, the cumulative error at each transformer block's output increases monotonically with depth. For dense geometric tasks, these amplified distortions lead to significant structural drift in the inferred 3D space. To mitigate this, we propose \textbf{TRE-Guided Module-wise Compensation}, a selective, low-rank correction mechanism.

\textbf{Low-rank Residual Compensation.} 
We apply compensation to the two core residual submodules in each transformer block: the Attention layer and the MLP layer. 
To rectify the residual error between the full-precision output $\mathbf{y}$ and its quantized counterpart $\mathbf{y}_{\Delta}$, we introduce a parallel lightweight branch that is inspired by QwT~\cite{fu2025quantization}. 
Furthermore, inspired by the observation that quantization error often exhibits low-rank characteristics~\cite{li2023loftq}, we formulate the compensated output $\hat{\mathbf{y}}_q$ as:
\begin{equation}
    \hat{\mathbf{y}}_q = \mathbf{y}_q + \mathbf{U}\mathbf{V}\mathbf{x} + b,
\end{equation}
where $\mathbf{x}$ is the submodule input, and $\mathbf{U} \in \mathbb{R}^{d \times r}, \mathbf{V} \in \mathbb{R}^{r \times d}$ constitute a low-rank adapter with rank $r \ll d$. The optimal compensation matrix $\mathbf{W}_{comp} \approx \mathbf{U}\mathbf{V}$ and bias $b$ is estimated via closed-form linear regression on the calibration set:
\begin{equation}
    (\mathbf{W}_{comp},b) = \mathop{\arg\min}_{\mathbf{W},b} \| \mathbf{y} - (\mathbf{y}_q + \mathbf{W}\mathbf{x} + b) \|_2^2.
\end{equation}
To minimize deployment overhead, we apply truncated Singular Value Decomposition (SVD) to $\mathbf{W}_{comp}$ and retain only the $r$ dominant singular components to construct $\mathbf{U}$ and $\mathbf{V}$. This sequential estimation ensures that each module's compensation accounts for errors propagated from all preceding layers.

\textbf{Tail Relative Error (TRE) Gating.} 
While module-wise compensation rectifies local bias, indiscriminate application to all layers incurs unnecessary parameter overhead and risks over-correction. Crucially, as shown in Fig.~\ref{fig:activation_error_tail}, quantization distortions are not uniformly distributed but are often dominated by high-magnitude activation outliers. Conventional global metrics like MSE fail to capture this sensitivity as they dilute the impact of outliers across the entire feature dimension.

We therefore introduce \textbf{Tail Relative Error (TRE)} to identify modules where high-magnitude distortions significantly compromise geometric fidelity. For a module output $\mathbf{y} \in \mathbb{R}^n$, we define the tail index set $\mathcal{S}$ containing the indices of the top-$k$ elements with the largest absolute magnitudes, where $k = \lfloor \rho n \rfloor$ (typically $\rho = 0.01$). The TRE is defined as:
\begin{equation}
    \text{TRE}(\mathbf{y}, \mathbf{y}_q) = \frac{\sum_{i \in \mathcal{S}} (y_i - y_{q,i})^2}{\sum_{i \in \mathcal{S}} y_i^2 + \epsilon},
\end{equation}
where $\epsilon$ is a small constant for numerical stability. TRE explicitly quantifies the relative energy distortion concentrated in the long-tailed regions of the activation distribution. We employ TRE as an adaptive gating mechanism: the compensation branch $(\mathbf{U}, \mathbf{V})$ is activated only if $\text{TRE} > \tau$, where $\tau$ is a predefined threshold. This selective activation ensures that hardware resources are prioritized for the most error-sensitive modules, preserving geometric consistency with minimal computational footprint.

\section{Experiments}

\begin{table}[t]
\caption{\textbf{Quantization results on the ETH3D dataset.}
Best results in each column are highlighted in \textbf{bold}.
Our method is highlighted in \colorbox{yellow!15}{light yellow}.}
\label{tab:merged_eth3d_quant_highlighted_fix}
\centering
\scriptsize
\setlength{\tabcolsep}{3pt}
\renewcommand{\arraystretch}{1.05}
\begin{tabular}{c| c| l| c c c c c c}
\toprule
\multirow{2}{*}{Model} & \multirow{2}{*}{Bit-Width} & \multirow{2}{*}{Method}
& \multicolumn{2}{c}{Acc. $\downarrow$}
& \multicolumn{2}{c}{Comp. $\downarrow$}
& \multicolumn{2}{c}{N.C. $\uparrow$} 
\\
\cmidrule(lr){4-5}
\cmidrule(lr){6-7}
\cmidrule(lr){8-9}
& & & Mean & Med. & Mean & Med. & Mean & Med. \\
\midrule
% ================= VGGT Model =================
\multirow{12}{*}{VGGT} 
& \multirow{6}{*}{W4A8}
  & RTN     & 0.879 & 0.868 & 3.504 & 2.827 & 0.543 & 0.568 \\
& & PTQ4ViT & 0.669 & 0.559 & 1.107 & 0.770 & 0.684 & 0.765 \\
& & ERQ     & 1.010 & 0.954 & 2.368 & 1.978 & 0.547 & 0.582 \\
& & RepQ    & 1.144 & 1.112 & 2.895 & 2.311 & 0.535 & 0.548 \\
& & GPTQ    & 0.872 & 0.834 & 2.686 & 2.046 & 0.539 & 0.563 \\
& & \gc Ours    & \gc \textbf{0.492} & \gc \textbf{0.367} & \gc \textbf{0.677} & \gc \textbf{0.446} & \gc \textbf{0.761} & \gc \textbf{0.876} \\
\cmidrule{2-9}
& \multirow{6}{*}{W6A6}
  & RTN     & 0.900 & 0.804 & 1.965 & 1.350 & 0.572 & 0.611 \\
& & PTQ4ViT & 0.389 & 0.254 & 0.440 & 0.276 & 0.806 & 0.914 \\
& & ERQ     & 0.598 & 0.455 & 0.679 & 0.358 & 0.737 & 0.839 \\
& & RepQ    & 0.946 & 0.855 & 2.222 & 1.547 & 0.577 & 0.612 \\
& & GPTQ    & 0.407 & 0.290 & 0.549 & 0.345 & 0.783 & 0.889 \\
& & \gc Ours    & \gc \textbf{0.334} & \gc \textbf{0.216} & \gc \textbf{0.406} & \gc \textbf{0.226} & \gc \textbf{0.813} & \gc \textbf{0.925} \\
\midrule
% ================= Pi3 Model =================
\multirow{12}{*}{Pi3} 
& \multirow{6}{*}{W4A8}
  & RTN     & 1.181 & 1.178 & 2.281 & 1.669 & 0.522 & 0.534 \\
& & PTQ4ViT & 0.413 & 0.289 & 0.480 & \textbf{0.307} & 0.780 & 0.898 \\
& & RepQ    & 0.976 & 0.940 & 3.062 & 2.514 & 0.513 & 0.514 \\
& & ERQ     & 0.869 & 0.833 & 2.802 & 2.173 & 0.559 & 0.585 \\
& & GPTQ    & 0.872 & 0.834 & 2.686 & 2.046 & 0.539 & 0.563 \\
& & \gc Ours    & \gc \textbf{0.392} & \gc \textbf{0.271} & \gc \textbf{0.478} & \gc 0.308 & \gc \textbf{0.781} & \gc \textbf{0.907} \\
\cmidrule{2-9}
& \multirow{6}{*}{W6A6}
  & RTN     & 1.134 & 1.135 & 2.829 & 2.095 & 0.525 & 0.534 \\
& & PTQ4ViT & 0.221 & 0.140 & 0.237 & 0.136 & 0.845 & 0.955 \\
& & RepQ    & 0.888 & 0.844 & 2.398 & 1.853 & 0.532 & 0.548 \\
& & ERQ     & 0.293 & 0.202 & 0.270 & 0.141 & 0.743 & 0.858 \\
& & GPTQ    & 0.407 & 0.290 & 0.549 & 0.345 & 0.783 & 0.889 \\
& & \gc Ours    & \gc \textbf{0.209} & \gc \textbf{0.130} & \gc \textbf{0.228} & \gc \textbf{0.131} & \gc \textbf{0.856} & \gc \textbf{0.960} \\
\bottomrule
\end{tabular}
\end{table}

\subsection{Experiment and Evaluation Settings}

\paragraph{Base Model.}
We evaluate VGGT-1B~\cite{wang2025vggt} and Pi3~\cite{wang2025pi}, and quantize each model directly from its pre-trained full-precision checkpoint without retraining.

\paragraph{Tasks and Datasets.}
We mainly evaluate point cloud reconstruction. For calibration, we randomly sample 20 instances from the DTU training split and construct an 8-sample calibration set using our Progressive Calibration Set Construction; this set is used for all experiments.
For evaluation, we report reconstruction results on 7Scenes and ETH3D.
We also evaluate the prediction of camera parameters on Co3Dv2 to test cross-task robustness.

\paragraph{Quantization Settings.}
We consider three common low-bit configurations: W4A8, W6A6, and W8A8.

\begin{table}[t]
\caption{\textbf{Quantization results on the 7Scenes dataset.}
Best results in each column are highlighted in \textbf{bold}.
Our method is highlighted in \colorbox{yellow!15}{light yellow}.}
\label{tab:merged_7scenes_quant_highlighted_fix}
\centering
\scriptsize
\setlength{\tabcolsep}{3pt}
\renewcommand{\arraystretch}{1.05}
\begin{tabular}{c| c| l| c c c c c c}
\toprule
\multirow{2}{*}{Model} & \multirow{2}{*}{Bit-Width} & \multirow{2}{*}{Method}
& \multicolumn{2}{c}{Acc. $\downarrow$}
& \multicolumn{2}{c}{Comp. $\downarrow$}
& \multicolumn{2}{c}{N.C. $\uparrow$} 
\\
\cmidrule(lr){4-5}
\cmidrule(lr){6-7}
\cmidrule(lr){8-9}
& & & Mean & Med. & Mean & Med. & Mean & Med. \\
\midrule
% ================= VGGT Model =================
\multirow{12}{*}{VGGT} 
& \multirow{6}{*}{W4A8}
  & RTN     & 0.059 & 0.038 & 0.657 & 0.575 & 0.586 & 0.631 \\
& & PTQ4ViT & 0.046 & 0.022 & 0.056 & 0.022 & 0.680 & 0.776 \\
& & ERQ     & 0.077 & 0.044 & 0.371 & 0.273 & 0.610 & 0.664 \\
& & RepQ    & 0.078 & 0.051 & 0.425 & 0.334 & 0.611 & 0.666 \\
& & GPTQ    & 0.086 & 0.057 & 0.422 & 0.306 & 0.594 & 0.646 \\
& & \gc Ours    & \gc \textbf{0.028} & \gc \textbf{0.011} & \gc \textbf{0.044} & \gc \textbf{0.015} & \gc \textbf{0.687} & \gc \textbf{0.788} \\
\cmidrule{2-9}
& \multirow{6}{*}{W6A6}
  & RTN     & 0.093 & 0.055 & 0.165 & 0.070 & 0.637 & 0.708 \\
& & PTQ4ViT & 0.023 & 0.009 & 0.039 & 0.015 & 0.690 & 0.792 \\
& & ERQ     & 0.048 & 0.024 & 0.067 & 0.029 & 0.671 & 0.764 \\
& & RepQ    & 0.127 & 0.082 & 0.180 & 0.075 & 0.594 & 0.645 \\
& & GPTQ    & 0.022 & 0.009 & 0.038 & \textbf{0.014} & 0.689 & 0.792 \\
& & \gc Ours    & \gc \textbf{0.021} & \gc \textbf{0.009} & \gc \textbf{0.037} & \gc \textbf{0.014} & \gc \textbf{0.690} & \gc \textbf{0.793} \\
\midrule
% ================= Pi3 Model =================
\multirow{12}{*}{Pi3} 
& \multirow{6}{*}{W4A8}
  & RTN     & 0.043 & 0.033 & 0.712 & 0.643 & 0.497 & 0.496 \\ 
& & PTQ4ViT
            & 0.027 & \textbf{0.013}
            & \textbf{0.044} & \textbf{0.014}
            & \textbf{0.674} & \textbf{0.771} \\
& & RepQ    & 0.069 & 0.047 & 0.451 & 0.344 & 0.493 & 0.490 \\
& & ERQ     & 0.068 & 0.045 & 0.426 & 0.326 & 0.536 & 0.554 \\
& & GPTQ    & 0.086 & 0.057 & 0.422 & 0.306 & 0.594 & 0.646 \\
& & \gc Ours    & \gc \textbf{0.026} & \gc \textbf{0.013} & \gc 0.046 & \gc 0.016 & \gc 0.672 & \gc 0.768 \\
\cmidrule{2-9}
& \multirow{6}{*}{W6A6}
  & RTN     & 0.057 & 0.043 & 0.568 & 0.481 & 0.503 & 0.504 \\
& & PTQ4ViT & 0.018 & 0.008 & 0.022 & 0.009 & 0.683 & 0.785 \\
& & RepQ    & 0.096 & 0.060 & 0.213 & 0.087 & 0.508 & 0.511 \\
& & ERQ
            & 0.021 & 0.010
            & \textbf{0.017} & \textbf{0.006}
            & 0.564 & 0.598 \\
& & GPTQ    & 0.022 & 0.009 & 0.038 & 0.014 & \textbf{0.689} & \textbf{0.792} \\
& & \gc Ours    & \gc \textbf{0.017} & \gc \textbf{0.007} & \gc 0.022 & \gc 0.009 & \gc 0.681 & \gc 0.782 \\
\bottomrule
\end{tabular}
\end{table}

\paragraph{Baseline Methods.}
We compare with representative PTQ baselines, including Round-to-nearest (RTN), PTQ4ViT~\cite{yuan2022ptq4vit}, RepQ-ViT~\cite{li2023repq}, ERQ, and GPTQ~\cite{frantar2022gptq}.
For GPTQ, whose original implementation quantizes weights only, we adopt PT4ViT activation quantizer.

For a fair and controlled comparison, we evaluate all methods under a unified \emph{per-tensor} weight quantization setting.
Our method is inherently agnostic to the choice of quantization granularity and can be applied under both per-tensor and channel-wise settings, whereas several baseline methods are originally formulated with channel-wise weights.
Allowing channel-wise quantization for these baselines in the main comparison would introduce an inherent advantage due to the additional per-channel scaling flexibility.
We therefore standardize all baselines to the per-tensor setting, ensuring that the comparison primarily reflects differences in calibration and error-mitigation strategies rather than quantizer granularity.
For completeness, we additionally compare our method with channel-wise variants of these baselines in Appendix~\ref{appendix:channel-wise}.

\paragraph{Implementation Details.}
Unless otherwise specified, all methods use the same calibration data and inference settings.
Additional implementation details, hyperparameters, and calibration configurations are provided in Appendix~\ref{appendix:details}.

\paragraph{Metrics.}
For point cloud reconstruction, we report Acc. and Comp. (lower is better) and N.C. (higher is better). For camera parameter prediction on Co3Dv2, we report AUC at multiple thresholds (A@30/A@15/A@5/A@3; higher is better).

\subsection{Main Results}

\paragraph{Quantitative Comparison.}

We evaluate TAPTQ on point cloud reconstruction across 7Scenes and ETH3D under two quantization settings, W4A8 and W6A6.
Results are summarized in Tables~\ref{tab:merged_7scenes_quant_highlighted_fix} and~\ref{tab:merged_eth3d_quant_highlighted_fix}, with additional experiments on DTU reported in Appendix~\ref{appendix:dtu}.
Lower Acc. and Comp. indicate better reconstruction quality, while higher N.C. reflects improved normal consistency.

Overall, TAPTQ consistently improves reconstruction performance across benchmarks and architectures, with larger gains observed under aggressive quantization and on more challenging scenes.
On 7Scenes, TAPTQ achieves the best or competitive results on both VGGT and Pi3.
Notably, under W4A8, TAPTQ clearly outperforms existing PTQ baselines on VGGT, while under W6A6, where performance becomes saturated, it remains among the top-performing methods without degradation.
On ETH3D, TAPTQ delivers more pronounced improvements across both models, highlighting its robustness to complex geometry and severe quantization noise.

\begin{table}[t]
  \caption{\textbf{Camera parameter prediction results.}
  Best results in each column are highlighted in \textbf{bold}.
  Our method is highlighted in \colorbox{yellow!15}{light yellow}.}
  \label{tab:camera_pose_auc}
  \centering
  \abovespace
  \scriptsize
  \setlength{\tabcolsep}{3pt}
  \renewcommand{\arraystretch}{1.05}
  \begin{tabular}{c| l| cccc}
    \toprule
    Bit-Width & Method  
    & A@30$\uparrow$ & A@15$\uparrow$ & A@5$\uparrow$ & A@3$\uparrow$ \\
    \midrule
    \multirow{6}{*}{W4A8}
    & RTN     & 0.0041 & 0.0007 & 0.0000 & 0.0000 \\
    & PTQ4ViT & 0.7622 & 0.6151 & 0.2753 & 0.1363 \\
    & GPTQ    & 0.0334 & 0.0137 & 0.0010 & 0.0005 \\
    & ERQ     & 0.0202 & 0.0043 & 0.0006 & 0.0002 \\
    & REPQ    & 0.0063 & 0.0006 & 0.0000 & 0.0000 \\
    & \gc Ours
              & \gc \textbf{0.8099} & \gc \textbf{0.6848} & \gc \textbf{0.3975} & \gc \textbf{0.2484} \\
    \midrule
    \multirow{6}{*}{W8A8}
    & RTN     & 0.8998 & 0.8190 & 0.5914 & 0.4380 \\
    & PTQ4ViT & 0.9220 & 0.8592 & 0.6877 & 0.5635 \\
    & GPTQ    & 0.9225 & 0.8608 & 0.6907 & 0.5686 \\
    & ERQ     & 0.9014 & 0.8193 & 0.5596 & 0.3906 \\
    & REPQ    & 0.8524 & 0.7373 & 0.4379 & 0.2827 \\
    & \gc Ours
              & \gc \textbf{0.9252} & \gc \textbf{0.8620} & \gc \textbf{0.6933} & \gc \textbf{0.5723} \\
    \bottomrule
  \end{tabular}
  \belowspace
\end{table}

\paragraph{Qualitative Comparison.} 
Figure~\ref{fig:vis_points} compares the W6A6 reconstruction results produced by PTQ4ViT and TAPTQ on VGGT and Pi3.
Under the same quantization setting, TAPTQ better preserves geometric structures and surface continuity.
Additional qualitative comparisons are provided in the appendix.

\paragraph{Camera Parameter Prediction.}
Table~\ref{tab:camera_pose_auc} reports camera parameter prediction results under W4A8 and W8A8.
This task is highly sensitive to quantization. Under W4A8, most baselines collapse, while TAPTQ substantially improves AUC across thresholds.
Under W8A8, TAPTQ achieves the best performance.

\subsection{Ablation Study}

\paragraph{Effect of Calibration Set Construction.}
Table~\ref{tab:calib_ablation_7scenes} compares calibration set construction strategies on 7Scenes under W4A8, using the same initial pool of 20 DTU training samples.
\emph{All samples} uses all 20 samples as a high-cost reference; \emph{rand} reports multiple random draws to show variance; \emph{Stability score} filters unstable samples using Appendix~\ref{appendix:stability} and then performs $K{=}4$ \emph{k}-means with proportional sampling.

As shown in Table~\ref{tab:calib_ablation_7scenes}, random selection shows large variance under small budgets.
With only 8 samples, our method achieves the best overall performance and even surpasses the 20-sample baseline, which highlights the importance of filtering out noisy samples.

\begin{table}[t]
  \caption{\textbf{Ablation on calibration set construction.} Model: VGGT, dataset: 7Scenes, setting: W4A8.
  All methods select calibration samples from the same initial pool of 20 DTU training samples.
  Best results in each column are highlighted in \textbf{bold}.
  Our method is highlighted in \colorbox{yellow!15}{light yellow}.}
  \label{tab:calib_ablation_7scenes}
  \centering
  \abovespace
  \scriptsize
  \setlength{\tabcolsep}{3pt}
  \renewcommand{\arraystretch}{1.05}
  \begin{tabular}{c l c c c c c c}
    \toprule
    Size & Method
    & \multicolumn{2}{c}{Acc.$\downarrow$}
    & \multicolumn{2}{c}{Comp.$\downarrow$}
    & \multicolumn{2}{c}{N.C.$\uparrow$} \\
    \cmidrule(lr){3-4}
    \cmidrule(lr){5-6}
    \cmidrule(lr){7-8}
    & & Mean & Med.
    & Mean & Med.
    & Mean & Med. \\
    \midrule

    \multirow{1}{*}{20}
    & All samples
    & 0.0281 & 0.0123
    & 0.0452 & 0.0161
    & 0.6864 & 0.7873 \\
    \midrule

    \multirow{8}{*}{4}
    & rand (case1)
    & 0.0370 & 0.0146
    & 0.0580 & 0.0241
    & 0.6842 & 0.7835 \\
    & rand (case2)
    & 0.0354 & 0.0158
    & 0.0495 & 0.0198
    & 0.6864 & 0.7863 \\
    & rand (case3)
    & 0.0310 & 0.0129
    & 0.0481 & 0.0163
    & \textbf{0.6874} & \textbf{0.7898} \\
    & rand (case4)
    & 0.0298 & 0.0117
    & 0.0456 & 0.0160
    & 0.6825 & 0.7823 \\
    & rand (case5)
    & 0.0332 & 0.0134
    & 0.0508 & 0.0178
    & 0.6851 & 0.7856 \\
    & rand (case6)
    & 0.0307 & 0.0122
    & 0.0468 & 0.0165
    & 0.6836 & 0.7819 \\
    \cmidrule(lr){2-8}
    & Stability Score
    & 0.0291 & 0.0122
    & 0.0470 & 0.0161
    & 0.6851 & 0.7852 \\
    & \gc Ours
    & \gc 0.0295 & \gc 0.0121
    & \gc 0.0460 & \gc 0.0166
    & \gc 0.6868 & \gc 0.7875 \\
    \midrule

    \multirow{5}{*}{8}
    & rand (case1)
    & 0.0300 & 0.0131
    & 0.0496 & 0.0177
    & 0.6868 & 0.7884 \\
    & rand (case2)
    & 0.0342 & 0.0139
    & 0.0449 & 0.0158
    & 0.6818 & 0.7806 \\
    & rand (case3)
    & 0.0318 & 0.0134
    & 0.0478 & 0.0169
    & 0.6842 & 0.7841 \\
    \cmidrule(lr){2-8}
    & Stability Score
    & 0.0330 & 0.0137 & \textbf{0.0436} & 0.0164 & 0.6811 & 0.7805 \\ 
    & \gc Ours
    & \gc \textbf{0.0264} & \gc \textbf{0.0105}
    & \gc 0.0444 & \gc \textbf{0.0147}
    & \gc 0.6867 & \gc 0.7884 \\
    \bottomrule
  \end{tabular}
  \belowspace
\end{table}

\paragraph{Effect of the Compensation Threshold $\tau$.}

\begin{table}[t]
  \caption{\textbf{Ablation on compensation threshold $\tau$ under W4A8.}
  Results on 7Scenes.}
  \label{tab:tau_ablation}
  \centering
  \abovespace
  \begin{small}
  \setlength{\tabcolsep}{4pt}
  \renewcommand{\arraystretch}{1.0}
  \resizebox{\linewidth}{!}{
  \begin{tabular}{l|cc cc cc c}
    \toprule
    Strategy
    & \multicolumn{2}{c}{Acc.$\downarrow$}
    & \multicolumn{2}{c}{Comp.$\downarrow$}
    & \multicolumn{2}{c}{N.C.$\uparrow$}
    & Param. (MB) \\
    \cmidrule(lr){2-3}
    \cmidrule(lr){4-5}
    \cmidrule(lr){6-7}
    & Mean & Med.
    & Mean & Med.
    & Mean & Med.
    &  \\
    \midrule
    No comp
    & 0.0457 & 0.0218
    & 0.0562 & 0.0222
    & 0.6795 & 0.7757
    & 608.0799 \\
    Full comp
    & 0.0291 & 0.0117
    & \textbf{0.0377} & 0.0147
    & 0.6863 & 0.7876
    & 616.9549 \\
    $\tau{=}0.005$
    & 0.0271 & 0.0110
    & 0.0392 & 0.0146
    & 0.6835 & 0.7842
    & 612.2049 \\
    $\tau{=}0.01$
    & 0.0279 & 0.0110
    & 0.0436 & 0.0153
    & 0.6868 & 0.7883
    & 610.3299 \\
    $\tau{=}0.02$
    & 0.0306 & 0.0125
    & 0.0489 & 0.0157
    & 0.6872 & 0.7883
    & 609.3924 \\
    $\tau{=}0.007$
    & \textbf{0.0251} & \textbf{0.0102}
    & 0.0401 & \textbf{0.0146}
    & \textbf{0.6878} & \textbf{0.7903}
    & 611.3299 \\
    \bottomrule
  \end{tabular}
  }
  \end{small}
  \belowspace
\end{table}

We further investigate the impact of the compensation threshold $\tau$ in the proposed TRE-Guided Compensation Module. Table~\ref{tab:tau_ablation} presents the results on 7Scenes under W4A8. 
As shown in Table~\ref{tab:tau_ablation}, quantization without compensation causes a substantial degradation in reconstruction quality. Full compensation recovers performance but may over-correct quantization-insensitive modules. In contrast, TRE-guided compensation provides a better performance by compensating only high-TRE modules. As $\tau$ increases from 0.005 to 0.02, the number of compensated modules decreases; $\tau{=}0.007$ achieves the best overall performance.

\paragraph{Effect of Ternary Search and Compensation.}
Table~\ref{tab:ternary_comp_ablation_7scenes} evaluates ternary search and module-wise compensation under W6A6.
Ternary search replaces exhaustive interval enumeration, substantially reducing calibration time with comparable accuracy.
Module-wise compensation further improves reconstruction quality with marginal overhead, indicating that the two components are complementary.

\begin{figure}[t]
  \centering
  \includegraphics[width=\columnwidth]{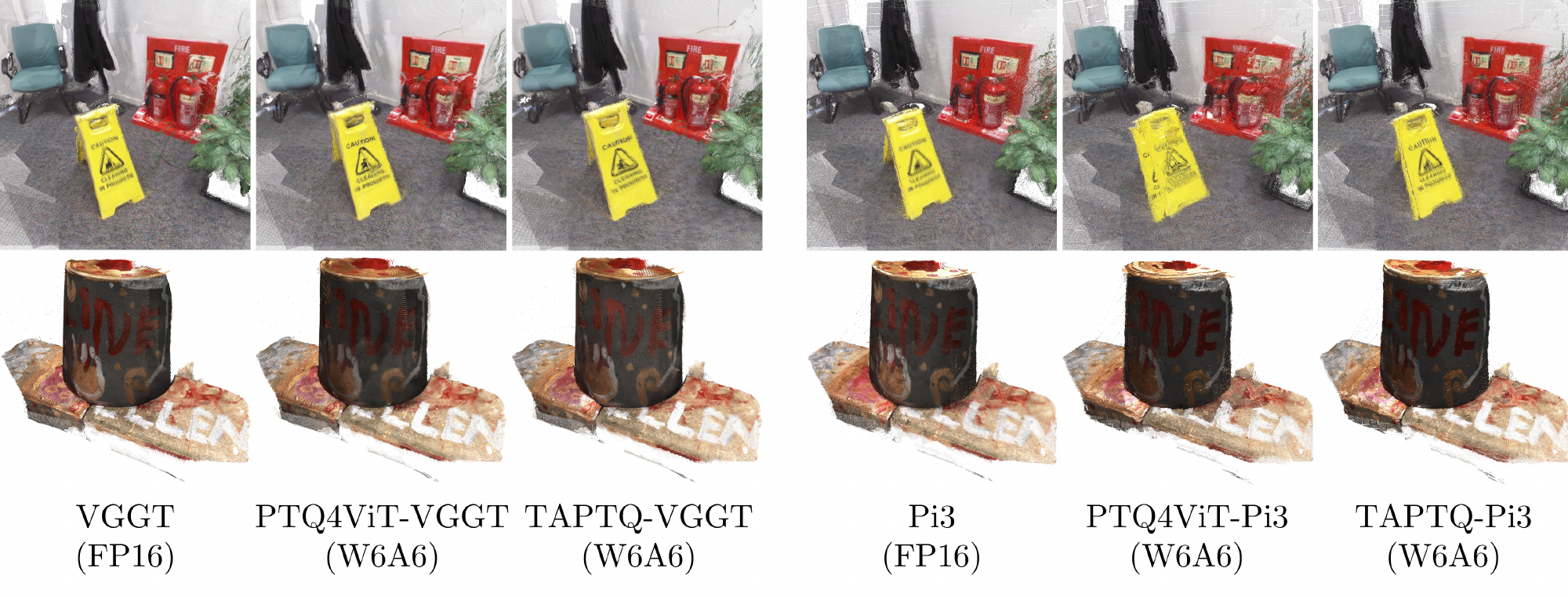}
  \caption{\textbf{Qualitative Comparison.}
    Visual comparison of W6A6 quantized reconstructions produced by PTQ4ViT and TAPTQ on VGGT and Pi3.
  }
  \label{fig:vis_points}
\end{figure}

\begin{table}[t]
  \caption{
    \textbf{Effect of ternary search and module-wise compensation under W6A6.}
    Quantitative results on the 7Scenes dataset that evaluate the reconstruction performance and calibration efficiency.
    All results report mean values; Time denotes calibration time.
    }

  \label{tab:ternary_comp_ablation_7scenes}
  \centering
  \scriptsize
  \setlength{\tabcolsep}{4pt}
  \renewcommand{\arraystretch}{1.05}
  \begin{tabular}{l c c c c c}
    \toprule
    \makecell[c]{Ternary \\ Search} & \makecell[c]{Error\\Compens.} 
    & Acc$\downarrow$
    & Comp$\downarrow$
    & N.C.$\uparrow$
    & Time\\
    \midrule
    \checkmark  & \checkmark
        & 0.021
        & 0.035
        & 0.692
        & 168.7\,min\\
    \checkmark  & \xmark
        & 0.023
        & 0.040
        & 0.688
        & 161.5\,min\\
    
    \midrule
    \xmark & \checkmark
        & 0.020
        & 0.036
        & 0.691
        & 529.7\,min\\
    \xmark & \xmark
        & 0.023
        & 0.039
        & 0.690
        & 522.6\,min\\
    
    \bottomrule
  \end{tabular}
\end{table}

\section{Conclusion}

We propose \textbf{TAPTQ}, a Tail-Aware Post-Training Quantization pipeline tailored for 3D geometry models.
TAPTQ combines \textbf{Progressive Calibration Set Construction} for stable range estimation under limited calibration budgets, \textbf{Quantization Interval Optimization via Ternary Search} to reduce interval calibration cost, and \textbf{TRE-Guided Module-wise Compensation} to selectively correct cross-layer error accumulation.
Across point cloud reconstruction benchmarks (7Scenes/ETH3D) and camera parameter prediction on Co3Dv2, TAPTQ consistently improves quantized performance under low-bit settings (e.g., W4A8/W6A6/W8A8), with particularly strong gains under aggressive quantization.
Ablations further validate the complementarity of ternary search and TRE-guided compensation, showing improved calibration efficiency and accuracy--consistency trade-offs.

\section*{Impact Statement}

This paper presents work whose goal is to advance the field of Machine
Learning. There are many potential societal consequences of our work, none
which we feel must be specifically highlighted here.

\bibliography{ref}

@inproceedings{wang2024dust3r,
  title={Dust3r: Geometric 3d vision made easy},
  author={Wang, Shuzhe and Leroy, Vincent and Cabon, Yohann and Chidlovskii, Boris and Revaud, Jerome},
  booktitle={Proceedings of the IEEE/CVF Conference on Computer Vision and Pattern Recognition},
  pages={20697--20709},
  year={2024}
}

@inproceedings{leroy2024grounding,
  title={Grounding image matching in 3d with mast3r},
  author={Leroy, Vincent and Cabon, Yohann and Revaud, J{\'e}r{\^o}me},
  booktitle={European Conference on Computer Vision},
  pages={71--91},
  year={2024},
  organization={Springer}
}

@inproceedings{wang2025vggt,
  title={Vggt: Visual geometry grounded transformer},
  author={Wang, Jianyuan and Chen, Minghao and Karaev, Nikita and Vedaldi, Andrea and Rupprecht, Christian and Novotny, David},
  booktitle={Proceedings of the Computer Vision and Pattern Recognition Conference},
  pages={5294--5306},
  year={2025}
}

@article{wang2025pi,
  title={$\pi^{3}$: Permutation-Equivariant Visual Geometry Learning},
  author={Wang, Yifan and Zhou, Jianjun and Zhu, Haoyi and Chang, Wenzheng and Zhou, Yang and Li, Zizun and Chen, Junyi and Pang, Jiangmiao and Shen, Chunhua and He, Tong},
  journal={arXiv preprint arXiv:2507.13347},
  year={2025}
}

@inproceedings{li2023q,
  title={Q-diffusion: Quantizing diffusion models},
  author={Li, Xiuyu and Liu, Yijiang and Lian, Long and Yang, Huanrui and Dong, Zhen and Kang, Daniel and Zhang, Shanghang and Keutzer, Kurt},
  booktitle={Proceedings of the IEEE/CVF International Conference on Computer Vision},
  pages={17535--17545},
  year={2023}
}

@article{lin2025depth,
  title={Depth anything 3: Recovering the visual space from any views},
  author={Lin, Haotong and Chen, Sili and Liew, Junhao and Chen, Donny Y and Li, Zhenyu and Shi, Guang and Feng, Jiashi and Kang, Bingyi},
  journal={arXiv preprint arXiv:2511.10647},
  year={2025}
}

@inproceedings{cui2017hsfm,
  title={HSfM: Hybrid structure-from-motion},
  author={Cui, Hainan and Gao, Xiang and Shen, Shuhan and Hu, Zhanyi},
  booktitle={Proceedings of the IEEE conference on computer vision and pattern recognition},
  pages={1212--1221},
  year={2017}
}

@inproceedings{tang2022mixed,
  title={Mixed-precision neural network quantization via learned layer-wise importance},
  author={Tang, Chen and Ouyang, Kai and Wang, Zhi and Zhu, Yifei and Ji, Wen and Wang, Yaowei and Zhu, Wenwu},
  booktitle={European conference on computer vision},
  pages={259--275},
  year={2022},
  organization={Springer}
}

@article{frantar2022gptq,
  title={Gptq: Accurate post-training quantization for generative pre-trained transformers},
  author={Frantar, Elias and Ashkboos, Saleh and Hoefler, Torsten and Alistarh, Dan},
  journal={arXiv preprint arXiv:2210.17323},
  year={2022}
}

@inproceedings{xiao2023smoothquant,
  title={Smoothquant: Accurate and efficient post-training quantization for large language models},
  author={Xiao, Guangxuan and Lin, Ji and Seznec, Mickael and Wu, Hao and Demouth, Julien and Han, Song},
  booktitle={International conference on machine learning},
  pages={38087--38099},
  year={2023},
  organization={PMLR}
}

@inproceedings{yuan2022ptq4vit,
  title={Ptq4vit: Post-training quantization for vision transformers with twin uniform quantization},
  author={Yuan, Zhihang and Xue, Chenhao and Chen, Yiqi and Wu, Qiang and Sun, Guangyu},
  booktitle={European conference on computer vision},
  pages={191--207},
  year={2022},
  organization={Springer}
}

@inproceedings{li2023repq,
  title={Repq-vit: Scale reparameterization for post-training quantization of vision transformers},
  author={Li, Zhikai and Xiao, Junrui and Yang, Lianwei and Gu, Qingyi},
  booktitle={Proceedings of the IEEE/CVF International Conference on Computer Vision},
  pages={17227--17236},
  year={2023}
}

@article{feng2025quantized,
  title={Quantized Visual Geometry Grounded Transformer},
  author={Feng, Weilun and Qin, Haotong and Wu, Mingqiang and Yang, Chuanguang and Li, Yuqi and Li, Xiangqi and An, Zhulin and Huang, Libo and Zhang, Yulun and Magno, Michele and others},
  journal={arXiv preprint arXiv:2509.21302},
  year={2025}
}

@inproceedings{nagel2020up,
  title={Up or down? adaptive rounding for post-training quantization},
  author={Nagel, Markus and Amjad, Rana Ali and Van Baalen, Mart and Louizos, Christos and Blankevoort, Tijmen},
  booktitle={International conference on machine learning},
  pages={7197--7206},
  year={2020},
  organization={PMLR}
}

@article{li2021brecq,
  title={Brecq: Pushing the limit of post-training quantization by block reconstruction},
  author={Li, Yuhang and Gong, Ruihao and Tan, Xu and Yang, Yang and Hu, Peng and Zhang, Qi and Yu, Fengwei and Wang, Wei and Gu, Shi},
  journal={arXiv preprint arXiv:2102.05426},
  year={2021}
}

@article{wei2022qdrop,
  title={Qdrop: Randomly dropping quantization for extremely low-bit post-training quantization},
  author={Wei, Xiuying and Gong, Ruihao and Li, Yuhang and Liu, Xianglong and Yu, Fengwei},
  journal={arXiv preprint arXiv:2203.05740},
  year={2022}
}

@article{he2025preserving,
  title={Preserving LLM Capabilities through Calibration Data Curation: From Analysis to Optimization},
  author={He, Bowei and Yin, Lihao and Zhen, Huiling and Liu, Shuqi and Wu, Han and Zhang, Xiaokun and Yuan, Mingxuan and Ma, Chen},
  journal={arXiv preprint arXiv:2510.10618},
  year={2025}
}

@inproceedings{schonberger2016structure,
  title={Structure-from-motion revisited},
  author={Schonberger, Johannes L and Frahm, Jan-Michael},
  booktitle={Proceedings of the IEEE conference on computer vision and pattern recognition},
  pages={4104--4113},
  year={2016}
}

@inproceedings{yao2018mvsnet,
  title={Mvsnet: Depth inference for unstructured multi-view stereo},
  author={Yao, Yao and Luo, Zixin and Li, Shiwei and Fang, Tian and Quan, Long},
  booktitle={Proceedings of the European conference on computer vision (ECCV)},
  pages={767--783},
  year={2018}
}

@article{mildenhall2021nerf,
  title={Nerf: Representing scenes as neural radiance fields for view synthesis},
  author={Mildenhall, Ben and Srinivasan, Pratul P and Tancik, Matthew and Barron, Jonathan T and Ramamoorthi, Ravi and Ng, Ren},
  journal={Communications of the ACM},
  volume={65},
  number={1},
  pages={99--106},
  year={2021},
  publisher={ACM New York, NY, USA}
}

@inproceedings{barron2021mip,
  title={Mip-nerf: A multiscale representation for anti-aliasing neural radiance fields},
  author={Barron, Jonathan T and Mildenhall, Ben and Tancik, Matthew and Hedman, Peter and Martin-Brualla, Ricardo and Srinivasan, Pratul P},
  booktitle={Proceedings of the IEEE/CVF international conference on computer vision},
  pages={5855--5864},
  year={2021}
}

@article{kerbl20233d,
  title={3D Gaussian splatting for real-time radiance field rendering.},
  author={Kerbl, Bernhard and Kopanas, Georgios and Leimk{\"u}hler, Thomas and Drettakis, George},
  journal={ACM Trans. Graph.},
  volume={42},
  number={4},
  pages={139--1},
  year={2023}
}

@article{wang2025faster,
  title={Faster vggt with block-sparse global attention},
  author={Wang, Chung-Shien Brian and Schmidt, Christian and Piekenbrinck, Jens and Leibe, Bastian},
  journal={arXiv preprint arXiv:2509.07120},
  year={2025}
}

@article{shen2025fastvggt,
  title={Fastvggt: Training-free acceleration of visual geometry transformer},
  author={Shen, You and Zhang, Zhipeng and Qu, Yansong and Zheng, Xiawu and Ji, Jiayi and Zhang, Shengchuan and Cao, Liujuan},
  journal={arXiv preprint arXiv:2509.02560},
  year={2025}
}

@incollection{gholami2022survey,
  title={A survey of quantization methods for efficient neural network inference},
  author={Gholami, Amir and Kim, Sehoon and Dong, Zhen and Yao, Zhewei and Mahoney, Michael W and Keutzer, Kurt},
  booktitle={Low-power computer vision},
  pages={291--326},
  year={2022},
  publisher={Chapman and Hall/CRC}
}

@article{banner2019post,
  title={Post training 4-bit quantization of convolutional networks for rapid-deployment},
  author={Banner, Ron and Nahshan, Yury and Soudry, Daniel},
  journal={Advances in neural information processing systems},
  volume={32},
  year={2019}
}

@inproceedings{nagel2019data,
  title={Data-free quantization through weight equalization and bias correction},
  author={Nagel, Markus and Baalen, Mart van and Blankevoort, Tijmen and Welling, Max},
  booktitle={Proceedings of the IEEE/CVF international conference on computer vision},
  pages={1325--1334},
  year={2019}
}

@article{lin2024awq,
  title={Awq: Activation-aware weight quantization for on-device llm compression and acceleration},
  author={Lin, Ji and Tang, Jiaming and Tang, Haotian and Yang, Shang and Chen, Wei-Ming and Wang, Wei-Chen and Xiao, Guangxuan and Dang, Xingyu and Gan, Chuang and Han, Song},
  journal={Proceedings of machine learning and systems},
  volume={6},
  pages={87--100},
  year={2024}
}

@inproceedings{xie2024mesongs,
    title={MesonGS: Post-training Compression of 3D Gaussians via Efficient Attribute Transformation},
    author={Xie, Shuzhao and Zhang, Weixiang and Tang, Chen and Bai, Yunpeng and Lu, Rongwei and Ge, Shijia and Wang, Zhi},
    booktitle={European Conference on Computer Vision},
    year={2024},
    organization={Springer}
}

@inproceedings{xie2024sizegs,
    title={SizeGS: Size-aware Compression of 3D Gaussian Splatting via Mixed Integer Programming},
    author={Xie, Shuzhao and Liu, Jiahang and Zhang, Weixiang and Ge, Shijia and Pan, Sicheng and Tang, Chen and Bai, Yunpeng and Zhang, Cong and Fan, Xiaoyi and Wang, Zhi},
    booktitle={ACM MM},
    year={2025}
}

@inproceedings{fu2025quantization,
  title={Quantization without tears},
  author={Fu, Minghao and Yu, Hao and Shao, Jie and Zhou, Junjie and Zhu, Ke and Wu, Jianxin},
  booktitle={Proceedings of the Computer Vision and Pattern Recognition Conference},
  pages={4462--4472},
  year={2025}
}

@article{li2023loftq,
  title={Loftq: Lora-fine-tuning-aware quantization for large language models},
  author={Li, Yixiao and Yu, Yifan and Liang, Chen and He, Pengcheng and Karampatziakis, Nikos and Chen, Weizhu and Zhao, Tuo},
  journal={arXiv preprint arXiv:2310.08659},
  year={2023}
}
\bibliographystyle{icml2026}

%%%%%%%%%%%%%%%%%%%%%%%%%%%%%%%%%%%%%%%%%%%%%%%%%%%%%%%%%%%%%%%%%%%%%%%%%%%%%%%
%%%%%%%%%%%%%%%%%%%%%%%%%%%%%%%%%%%%%%%%%%%%%%%%%%%%%%%%%%%%%%%%%%%%%%%%%%%%%%%
% APPENDIX
%%%%%%%%%%%%%%%%%%%%%%%%%%%%%%%%%%%%%%%%%%%%%%%%%%%%%%%%%%%%%%%%%%%%%%%%%%%%%%%
%%%%%%%%%%%%%%%%%%%%%%%%%%%%%%%%%%%%%%%%%%%%%%%%%%%%%%%%%%%%%%%%%%%%%%%%%%%%%%%
\clearpage
\newpage
\appendix

\section{Appendix: Additional Details}

\subsection{Ternary-Search-Based Quantization Interval Search}
\label{appendix:Ternary}
\begin{algorithm}[h]
  \caption{Ternary-Search-Based Quantization Interval Search}
  \label{alg:ternary_interval_search}
  \begin{algorithmic}
    \STATE {\bfseries Input:} raw input $x_0$, 
    raw output $y_0$, 
    interval candidates  $\{\alpha_i\}_{i=1}^{N}$, 
    similarity metric $\mathrm{Sim}(\cdot)$, 
    layer function $\mathcal{F}(\cdot;\alpha)$, 
    minimum interval $\epsilon$
    \STATE {\bfseries Output:} optimal quantization interval $\alpha^{*}$
    \STATE $L = 1, R = N$
    \REPEAT
        \STATE $m_1 = L + (R - L)/3$
        \STATE $m_2 = R - (R - L)/3$
        \STATE $y_{m_1} = \mathcal{F}(x_0;\alpha_{m_1})$
        \STATE $y_{m_2} = \mathcal{F}(x_0;\alpha_{m_2})$
        \STATE $S_{m_1} = Sim(y_0, y_{m_1})$
        \STATE $S_{m_2} = Sim(y_0, y_{m_2})$
        \IF{$S_{m_1} < S_{m_2}$}
            \STATE $L = m_1$
        \ELSE
            \STATE $R = m_2$
        \ENDIF
    \UNTIL{$R - L \leq \epsilon$}
    \STATE $\alpha^{*} = \alpha_{(L+R)/2}$
  \end{algorithmic}
\end{algorithm}

\subsection{Stability Score}
\label{appendix:stability}

To quantify the effect of our Stage-1 coarse-grained outlier suppression, we follow the noise-filtering principle of QuantVGGT and compute a stability score that measures the inherent stability of individual samples through feature statistics (used only as an auxiliary diagnostic, not a component of TAPTQ).
Specifically, for a calibration sample $x_i$, let $\mathbf{T}_l(x_i) \in \mathbb{R}^{N \times C}$ denote the aggregated token representations at layer $l$.
We first compute the average token-wise feature variance at each layer as
\begin{equation}
    v_l(x_i) = \frac{1}{N} \sum_{n=1}^{N} 
    \mathrm{Var}\!\left( \mathbf{T}_l^{(n)}(x_i) \right),
\end{equation}
where $\mathbf{T}_l^{(n)}(x_i) \in \mathbb{R}^{C}$ denotes the feature vector of the $n$-th token.
The sample-level noise magnitude is then obtained by averaging across layers:
\begin{equation}
    \bar{v}(x_i) = \frac{1}{L} \sum_{l=1}^{L} v_l(x_i),
\end{equation}
and the stability score is defined as the inverse of the aggregated variance:
\begin{equation}
    s_i = \frac{1}{\bar{v}(x_i) + \epsilon},
\end{equation}
where $L$ denotes the number of layers considered and $\epsilon$ is a small constant for numerical stability.
Larger values of $s_i$ correspond to lower feature variance and indicate higher sample stability.

\begin{figure}[h]
  \centering
  \includegraphics[width=\columnwidth]{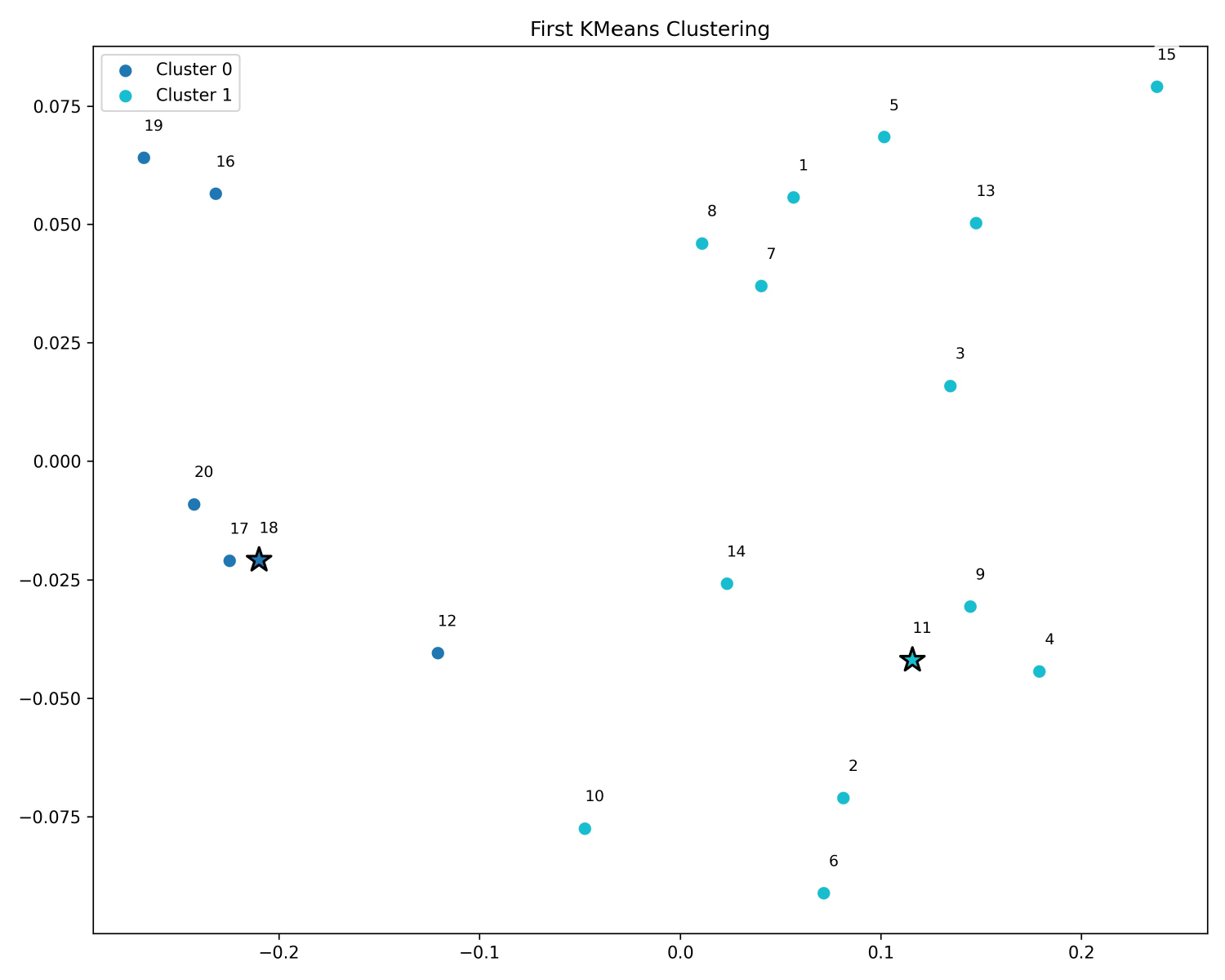}
    \caption{\textbf{First-stage clustering for noise separation.}
    Each point represents one calibration sample from the initial pool, where the overlaid number indicates its stability score rank (lower rank corresponds to higher stability). A two-cluster K-means is applied to separate unstable samples from relatively stable ones, which are retained for the second-stage clustering.}
  \label{fig:first_kmeans}
\end{figure}

As shown in Fig.~\ref{fig:first_kmeans}, when applied to a randomly selected pool of 20 samples, unstable samples are grouped together, indicating that this step effectively filters out noisy data.

\section{Appendix: Experimental Details}
\label{appendix:details}

\subsection{Calibration Data and Clustering Hyperparameters}
Following the main text, we sample 20 instances from the DTU training split as an initial pool, and construct an 8-sample calibration set using \textbf{Progressive Calibration Set Construction}.
For the clustering procedure, we use $K{=}2$ in Stage 1 and $K{=}N/2$ in Stage 2 (where $N$ denotes the target calibration set size).

\subsection{TRE-Guided Module-wise Compensation Hyperparameters}
We use tail ratio $\rho{=}0.01$ for \textbf{Tail Relative Error (TRE)} and set the compensation threshold to $\tau{=}0.007$ unless otherwise specified. For the low-rank compensation module, we set the rank to $r{=}16$.

\section{Appendix: Additional Discussion}
\label{appendix:discussion}

\subsection{Discussion on Calibration Set Selection vs. QuantVGGT}
QuantVGGT~\cite{feng2025quantized} highlights that multi-view calibration in 3D geometry models can be unstable, and proposes model-dependent strategies to filter and sample calibration data.
In contrast, our \textbf{Progressive Calibration Set Construction} adopts a \emph{model-agnostic} coarse-to-fine clustering strategy in a feature space, balancing outlier suppression and representativeness without relying on any model-specific internal signals; as a result, it is directly applicable across architectures (e.g., VGGT and Pi3).
We additionally follow QuantVGGT and report the Stability Score (Appendix~\ref{appendix:stability}) to quantify the effect of Stage-1 coarse-grained outlier suppression; TAPTQ itself remains model-independent in the calibration data selection stage.

\section{Appendix: More Experiments}
\label{more exp}

\subsection{Channel-wise Experiment}
\label{appendix:channel-wise}

\begin{table}[h] 
  \caption{\textbf{Channel-wise quantization results.} Results on the 7Scenes dataset using the VGGT model.
  Best results in each column are highlighted in \textbf{bold}.
  Our method is highlighted in \colorbox{yellow!15}{light yellow}.}
  \label{tab:vggt_7scenes_quant_single}
  \centering
  \scriptsize
  \setlength{\tabcolsep}{3pt}
  \renewcommand{\arraystretch}{1.05}
  \begin{tabular}{c| l| c c c c c c}
    \toprule
    Bit-Width & Method
    & \multicolumn{2}{c}{Acc.$\downarrow$}
    & \multicolumn{2}{c}{Comp.$\downarrow$}
    & \multicolumn{2}{c}{N.C.$\uparrow$} \\
    \cmidrule(lr){3-4}
    \cmidrule(lr){5-6}
    \cmidrule(lr){7-8}
    & &
    Mean & Med.
    & Mean & Med.
    & Mean & Med. \\
    \midrule
    \multirow{4}{*}{W4A8}
    & ERQ (ch-wise)  
      & 0.026 & 0.012 & 0.043 & 0.018 & 0.683 & 0.783 \\
    & RepQ (ch-wise)
      & 0.058 & 0.027 & 0.100 & 0.032 & 0.661 & 0.747 \\
    & GPTQ (ch-wise)
      & \textbf{0.021} & \textbf{0.009} & \textbf{0.033} & 0.017 & 0.686 & 0.788 \\
    & \gc Ours
      & \gc 0.028 & \gc 0.011 & \gc 0.044 & \gc \textbf{0.015} & \gc \textbf{0.687} & \gc \textbf{0.788} \\
    \midrule
    \multirow{4}{*}{W6A6}
    & ERQ (ch-wise)
      & 0.033 & 0.014 & 0.045 & 0.018 & 0.675 & 0.773 \\
    & RepQ (ch-wise)
      & 0.037 & 0.016 & 0.048 & 0.019 & 0.680 & 0.778 \\
    & GPTQ (ch-wise)
      & \textbf{0.021} & \textbf{0.009} & \textbf{0.036} & 0.014 & \textbf{0.691} & \textbf{0.794} \\
    & \gc Ours
      & \gc \textbf{0.021} & \gc \textbf{0.009} & \gc 0.037 & \gc \textbf{0.014} & \gc 0.690 & \gc 0.793 \\
    \bottomrule
  \end{tabular}
\end{table}

\subsection{Results on DTU}
\label{appendix:dtu}
% ------------------------------------------------------------
% ---------------------------VGGT------------------------------
% ------------------------------------------------------------

\begin{table}[h]
  \caption{\textbf{Quantization results on the DTU dataset.} Quantization results on the DTU dataset using the VGGT model.
  We report mean and median values. Best results in each column (within the same quantization setting) are highlighted in \textbf{bold}.
  Our method is highlighted in \colorbox{yellow!15}{light yellow}.}
  \label{tab:vggt_dtu_quant_vertical}
  \centering
  \scriptsize
  \setlength{\tabcolsep}{3pt}
  \renewcommand{\arraystretch}{1.05}
  \begin{tabular}{c| l| c c c c c c}
    \toprule
    Bit-Width & Method
    & \multicolumn{2}{c}{Acc.$\downarrow$}
    & \multicolumn{2}{c}{Comp.$\downarrow$}
    & \multicolumn{2}{c}{N.C.$\uparrow$} \\
    \cmidrule(lr){3-4}
    \cmidrule(lr){5-6}
    \cmidrule(lr){7-8}
    & &
    Mean & Med.
    & Mean & Med.
    & Mean & Med. \\
    \midrule
    \multirow{6}{*}{W4A8}
    & RTN
    & 8.093 & 6.168 & 32.707 & 20.993 & 0.586 & 0.629 \\
    & ERQ
    & 7.315 & 5.296 & 19.563 & 8.523 & 0.634 & 0.697 \\
    & RepQ
    & 10.011 & 7.740 & 20.115 & 7.943 & 0.610 & 0.662 \\
    & GPTQ
    & 8.117 & 5.973 & 32.042 & 20.285 & 0.604 & 0.651 \\
    & \gc Ours
    & \gc \textbf{1.147} & \gc \textbf{0.648} & \gc \textbf{2.179} & \gc \textbf{0.990} & \gc \textbf{0.682} & \gc \textbf{0.770} \\
    \midrule
    \multirow{6}{*}{W6A6}
    & RTN
    & 7.879 & 5.748 & 11.939 & 3.806 & 0.653 & 0.723 \\
    & ERQ
    & 1.229 & 0.704 & 2.178 & 1.144 & \textbf{0.684} & \textbf{0.769} \\
    & RepQ
    & 10.395 & 7.445 & 19.965 & 10.676 & 0.605 & 0.654 \\
    & GPTQ
    & 1.204 & 0.679 & 2.000 & 0.935 & 0.679 & 0.767 \\
    & \gc Ours
    & \gc \textbf{1.190} & \gc \textbf{0.671} & \gc \textbf{1.818} & \gc \textbf{0.908} & \gc 0.678 & \gc 0.765 \\
    \midrule
    \multirow{6}{*}{W8A8}
    & RTN
    & 1.216 & 0.729 & 2.174 & 1.192 & 0.687 & 0.772 \\
    & PTQ4ViT
    & \textbf{1.136} & \textbf{0.665} & 2.043 & \textbf{1.135} & 0.691 & 0.776 \\
    & ERQ
    & 1.196 & 0.715 & 2.189 & 1.256 & \textbf{0.693} & \textbf{0.778} \\
    & GPTQ
    & 1.166 & 0.686 & 2.081 & 1.182 & 0.692 & 0.777 \\
    & \gc Ours
    & \gc 1.153 & \gc 0.673 & \gc \textbf{2.025} & \gc \textbf{1.135} & \gc 0.692 & \gc 0.776 \\
    \bottomrule
  \end{tabular}
\end{table}

% ------------------------------------------------------------
% ---------------------------Pi3------------------------------
% ------------------------------------------------------------

\begin{table}[h]
  \caption{\textbf{Quantization results on the DTU dataset.} Quantization results on the DTU dataset using the Pi3 model.
  We report mean and median values. Best results in each column (within the same quantization setting) are highlighted in \textbf{bold}.
  Our method is highlighted in \colorbox{yellow!15}{light yellow}.}
  \label{tab:pi3_dtu_quant}
  \centering
  \scriptsize
  \setlength{\tabcolsep}{3pt}
  \renewcommand{\arraystretch}{1.05}
  \begin{tabular}{l| c| c c c c c c}
    \toprule
    Bit-Width & Method
    & \multicolumn{2}{c}{Acc.$\downarrow$}
    & \multicolumn{2}{c}{Comp.$\downarrow$}
    & \multicolumn{2}{c}{N.C.$\uparrow$} \\
    \cmidrule(lr){3-4}
    \cmidrule(lr){5-6}
    \cmidrule(lr){7-8}
    & & Mean & Med. & Mean & Med. & Mean & Med. \\
    \midrule
    % ================= W4A8 =================
    \multirow{6}{*}{W4A8}
    & RTN
      & 10.099 & 7.670
      & 17.295 & 4.441
      & 0.496 & 0.494 \\
    & PTQ4ViT & 2.259 & 1.301 & 2.346 & 0.740 & \textbf{0.682} & \textbf{0.768} \\
    & RepQ    & 8.769 & 6.654 & 15.075 & 2.680 & 0.501 & 0.501 \\
    & ERQ     & 8.437 & 6.272 & 18.061 & 5.838 & 0.532 & 0.550 \\
    & GPTQ    & 8.117 & 5.973 & 32.042 & 20.285 & 0.604 & 0.651 \\
    & \gc Ours
             & \gc \textbf{1.759} & \gc \textbf{0.960}
             & \gc \textbf{2.239} & \gc \textbf{0.663}
             & \gc 0.677 & \gc 0.765 \\
    \midrule
    % ================= W6A6 =================
    \multirow{6}{*}{W6A6}
    & RTN
      & 10.785 & 8.225
      & 18.575 & 5.908
      & 0.498 & 0.497 \\
    & PTQ4ViT & 2.557 & 1.496 & 2.174 & 0.769 & \textbf{0.681} & \textbf{0.767} \\
    & RepQ    & 8.327 & 6.037 & 14.617 & 3.167 & 0.517 & 0.526 \\
    & ERQ     & 1.710 & 0.900 & \textbf{1.623} & \textbf{0.566} & 0.540 & 0.561 \\
    & GPTQ
             & \textbf{1.204} & \textbf{0.679}
             & 2.000 & 0.935
             & 0.679 & 0.767 \\
    & \gc Ours    & \gc 1.576 & \gc 0.845 & \gc 1.999 & \gc 0.632 & \gc 0.669 & \gc 0.754 \\
    \midrule
    % ================= W8A8 =================
    \multirow{6}{*}{W8A8}
    & RTN
      & 1.721 & 0.933
      & 2.255 & 0.667
      & 0.676 & 0.763 \\
    & PTQ4ViT & 1.177 & 0.648 & 1.767 & 0.596 & 0.670 & 0.756 \\
    & RepQ    & 1.958 & 1.099 & 1.870 & 0.618 & 0.597 & 0.649 \\
    & ERQ     & 1.254 & 0.676 & \textbf{1.538} & \textbf{0.556} & 0.617 & 0.679 \\
    & GPTQ
             & 1.166 & 0.686
             & 2.081 & 1.182
             & \textbf{0.692} & \textbf{0.777} \\
    & \gc Ours
             & \gc \textbf{1.154} & \gc \textbf{0.637}
             & \gc 1.753 & \gc 0.605
             & \gc 0.669 & \gc 0.755 \\
    \bottomrule
  \end{tabular}
\end{table}

\subsection{Qualitative Comparison}

\begin{figure*}[t]
  \centering
  \includegraphics[width=\textwidth]{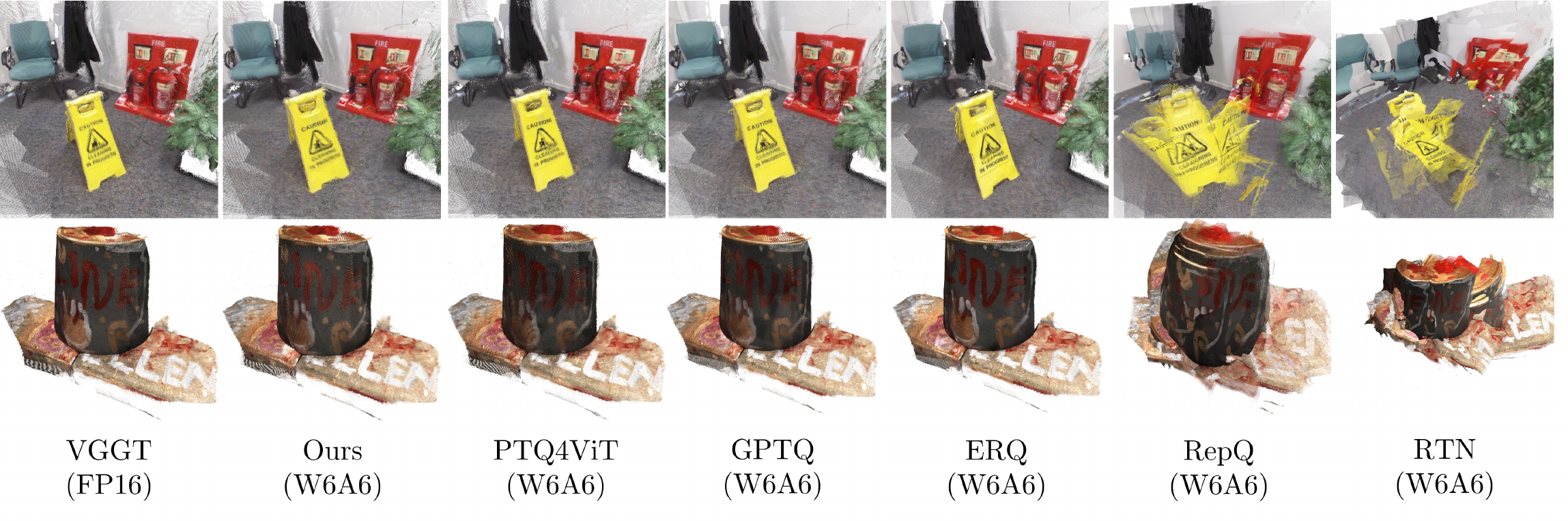}
  \caption{\textbf{Qualitative Comparison on VGGT.}
  }
  \label{fig:vis_points}
\end{figure*}

\begin{figure*}[t]
  \centering
  \includegraphics[width=\textwidth]{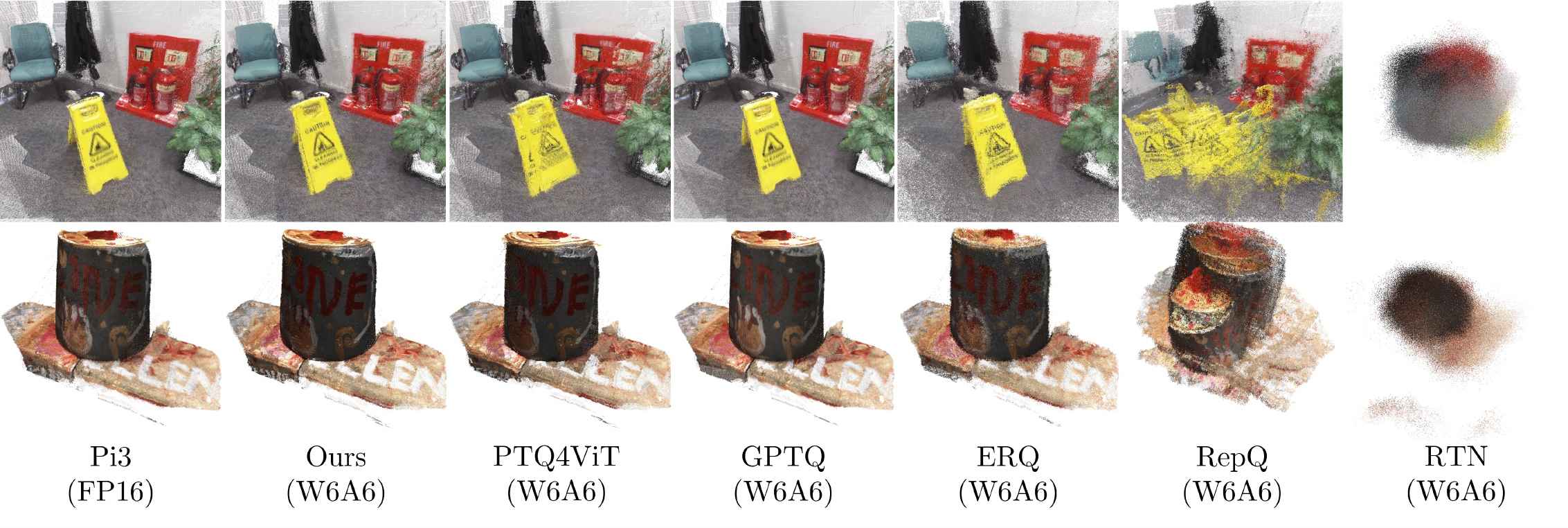}
  \caption{\textbf{Qualitative Comparison on Pi3.}
  }
  \label{fig:vis_points}
\end{figure*}

%%%%%%%%%%%%%%%%%%%%%%%%%%%%%%%%%%%%%%%%%%%%%%%%%%%%%%%%%%%%%%%%%%%%%%%%%%%%%%%
%%%%%%%%%%%%%%%%%%%%%%%%%%%%%%%%%%%%%%%%%%%%%%%%%%%%%%%%%%%%%%%%%%%%%%%%%%%%%%%

\end{document}